\documentclass[11pt]{article}
\usepackage[T1]{fontenc}
\usepackage{lmodern}
\usepackage[margin=1in]{geometry}
\usepackage{amsmath,amssymb}
\usepackage{graphicx}
\usepackage{booktabs}
\usepackage[numbers]{natbib}
\usepackage[hidelinks]{hyperref}
\graphicspath{{figures/}}

\newcommand{\Psync}{P_{\mathrm{sync}}}

\title{Degradation of Language Models under Controlled Load: Synchronization, Rule Execution and Misfires on the Background}
\author{Hennadii Kutomanov\\
  Independent researcher\\
  \texttt{kutomanov.hennadii@gmail.com}}
\date{}

\begin{document}
\maketitle
\begin{abstract}

As the load grows, a language model makes more and more errors, but different processes stand behind this decline in quality. The model may lose the correspondence with the source text, it may keep its place in the text but execute a rule wrongly, or it may change a word that no rule concerns. In this work the degradation of a model under controlled load is separated into these three components---synchronization, rule execution and misfires. The load is set by rules that replace one word with another in a meaningless text, and each kind of load---the text length, the position of a word in the answer, the replacement density, the number of rules and the place of a rule in the list---is set separately. The alignment of the answer with the reference makes it possible to follow the outcome of every word, and a statistical model that takes into account that the spread of difficulty between requests itself depends on the load restores, for each component, the laws of its dependence on every kind of load. These laws make up the passport of a model. The paper presents the method of the separation, the principles of constructing the laws and the passports of seven models built by this method; the separation opens the way to studying the fine structure of the behavior of a neural network.

\end{abstract}

\section{Introduction}\label{sec:introduction}

Language models increasingly face long requests containing dozens of instructions, yet as the load grows the quality of their execution declines. To study this degradation quantitatively one needs experimental tasks with a controlled load and an objective programmatic check of every action of the model. The previous work \citep{kutomanov2026capacity} developed such an instrument: the model receives a set of word replacement rules and a text made of random words unrelated in meaning. Its task is to reproduce the text, performing all the replacements prescribed by the rules. The text length, the share of words subject to replacement and the number of rules are set independently, and the correctness of every replacement is checked by a program. From the data obtained, quantitative regularities of the decline in instruction following were established, and the instruction capacity was defined---the limiting number of rules that the model is able to follow with a given share of correct replacements.

The previous work, however, evaluated only one outcome: whether the correct replacement stands in the place of the word to which a rule applies. Such a measure does not make it possible to determine the nature of the errors. The model may lose the correspondence with the source text by skipping a fragment, returning to lines already written or starting to repeat them. It may keep the right position but perform the replacement wrongly. Finally, the model may erroneously change a word to which no rule applies. The former measure mixed the first two kinds of errors and did not account for the third.

This work develops methods that make it possible to consider these three kinds of errors separately. Instead of a line-by-line comparison, the answer is aligned with the reference, which establishes the correspondence between individual words. On the basis of the alignment three quantities are defined: synchronization, which characterizes how the words of the answer keep their correspondence with their positions in the reference; rule execution, evaluated on the words subject to replacement that have kept the correspondence with their positions; and misfires, registered on background words of the same kind. A separate statistical model is built for each quantity. The dependences on the load axes are revealed by the profiles of its coefficients and then described by quantitative laws, from which the corresponding capacities are computed.

The measurements also showed that with the load not only the average quality of task execution changes, but also the spread of difficulty between individual requests. Therefore, instead of a GLMM with a constant variance of the random request intercept, a GLMM with heterogeneous random-intercept variance is used, in which the variance depends on the load axes. The random intercept characterizes the deviation of the difficulty of an individual request from the typical request, and its variance describes the size of the spread between requests. Taking the dependence of this variance on the load into account refines the laws of degradation and revises the capacity estimates derived from them. The heterogeneous variance is checked against the median request in the data: if it turns a law against the data, the quantity is fitted with one variance.

This work presents the passports of seven models from three developers: GPT-4o mini and GPT-4.1 mini by OpenAI, Gemini 3.5 Flash Lite and Gemini 3.1 Flash Lite by Google, Nemotron 3.5 Lightning, Nemotron 3 Nano Omni and Nemotron 3 Ultra by NVIDIA. All models are measured by one protocol, the passport of each of them describes three quantities, and this makes it possible for the first time to compare the laws of degradation of different models.

The contribution of this work comprises the following results:
\begingroup\sloppy
\begin{itemize}
\item a measure of synchronization and a method of aligning the answer with the reference by word positions are developed and verified on the published data of the previous work. The method separates the loss of correspondence of words with their positions from replacement errors and also detects erroneous replacements of background words (Sections~\ref{sec:three-quantities} and~\ref{sec:alignment});

\item three independent characteristics of the quality of task execution are distinguished---synchronization, rule execution and misfires. For each of them a separate statistical model is built and the laws of dependence on the load axes are established. The laws are fitted over the whole studied region, including the collapse boundary of the answer (Subsections~\ref{sec:quantities} and~\ref{sec:collapse-boundary}, Sections~\ref{sec:models} and~\ref{sec:passports});

\item a GLMM with heterogeneous random-intercept variance is developed, which accounts for the dependence of the spread of difficulty between individual requests on the load axes, together with a check of this model against the median request in the data. On its basis the laws of the influence of individual load factors are refined (Section~\ref{sec:spread});

\item principles of constructing smoothed laws are formulated: outliers are clipped at three standard deviations, points enter the fit with weights $1/(\sigma_i^2 + \tau^2)$, where $\tau^2$ is the spread of the points around the law beyond their uncertainties, and the form of a law is chosen by the smallest QAIC criterion without a ban on growth (Section~\ref{sec:smooth-laws});

\item the system of capacity measures is revised: the guaranteed capacities of the previous work are replaced by the measure of the density---a factor showing how many times the full replacement density multiplies the odds of an error compared with the rarest replacements; a factor below one means that the odds of an error decrease. The measure is defined for all three quantities (Subsection~\ref{sec:passports-density});

\item the level at which capacities are defined is changed from 0.95 to 0.5. With a heterogeneous variance, a capacity at the level 0.95 depends on the load at which the variance is evaluated, whereas at the level 0.5 this dependence is absent (Subsection~\ref{sec:spread-level});

\item the passports of seven models from three developers are built and compared. A passport includes the laws of dependence of the three quantities on the load axes, the capacities and the measure of the density. The models differ in their profile: the instruction capacity lies between 10 and 36 rules, while the fatigue capacities order the models differently. The models diverge most strongly in their response to the replacement density---the full density multiplies the odds of losing the place in the answer by a factor from 100 to $5.7\cdot 10^{12}$---and on part of the axes the very form of the laws differs between the models (Section~\ref{sec:passports});

\item version 0.2.3 of the open part of the experimental instrument is published, which makes it possible to recompute the numerical characteristics of the passports independently from the published experimental points; the passports of the three Nemotron models are published without points (Section~\ref{sec:availability}).
\end{itemize}
\endgroup

The value of the work lies in two interrelated results. Scientifically, it shows that the degradation of language models under load manifests itself in three different forms, each of which obeys its own regularities. The quantitative laws obtained, their parameters and uncertainty estimates make it possible to compare the behavior of different models. Practically, the passport of a model makes it possible to determine which factor limits the quality of task execution at a given load: the loss of synchronization with the source text, the wrong execution of a rule or a misfire.

The paper is organized as follows. Section~\ref{sec:related-work} relates this study to earlier works on the behavior of language models under load. Section~\ref{sec:three-quantities} motivates the transition from the single quality measure used in the previous work to three independent quantities. Section~\ref{sec:alignment} describes the method of aligning the answer with the reference and its verification, and Section~\ref{sec:models} the statistical models of the three quantities. Section~\ref{sec:spread} is devoted to modeling the heterogeneous random-intercept variance, its check against the median request and the justification of the level at which capacities are defined. Section~\ref{sec:smooth-laws} sets out the principles of constructing smoothed laws. Section~\ref{sec:passports} compares the passports of seven models. Section~\ref{sec:outlook} discusses the prospects for further research, and Section~\ref{sec:availability} gives information on the availability of the source code and the experimental data.

\section{Related Work}\label{sec:related-work}

The previous work \citep{kutomanov2026capacity}, which also contains a review of studies on the behavior of language models under load, measured the decline in instruction following as a quantity: it established the laws of its dependence on the kinds of load and derived capacities from them. This work continues that study and starts from the premise that degradation is not homogeneous: a model may lose its place in the answer, execute a rule wrongly or change a word that no rule concerns. Each of these forms has already been observed in other works, but separately and in different settings.

The loss of correspondence with the source text is observed both in applied tasks and in purpose-built experiments. \citet{liu2025llmigrate} studied the translation of programs from C to Rust and found that GPT-4o omitted substantial fragments of the source code. For excerpts shorter than 200 lines such omissions occurred in 1.67\% of cases, whereas for excerpts of about 700 lines their frequency reached 47.4\%. \citet{yang2025hallucinate} showed that in document summarization the factual accuracy of the last fifth of the answer is lower than that of the preceding parts: for Llama~3.1-8B it was 0.75 against 0.79--0.82. \citet{amit2026salt} built tasks with a single long answer that is generated by a program and checked without a judge model, and studied more than 50 models on them. The accuracy at the beginning of the answer turned out to be higher than at its end, and errors propagate from an already corrupted part of the answer: with 15\% of erroneous units in the preceding part, the share of correct subsequent units drops to about 50\%.

The loss of place is studied most directly in copying experiments. \citet{wen2026copy} showed that even frontier models have difficulty copying strings exactly when their length is far below the available context window, and the share of error-free copies decreases as the string length and the number of repeated substrings grow. The authors suggest that in copying the model relies not on the absolute position but on the similarity of the local context. \citet{zhang2026copygen} studied the verbatim reproduction of a given span of a contract, a scientific text, program code or a random sequence and distinguished two kinds of errors: a drift within the span, in which the correct token is replaced by a plausible one, and an overrun past the end of the span. Even the best of the frontier models studied, DeepSeek-V4-Pro, reproduced contract spans exactly in only 56.03\% of cases, and GPT-4.1-nano in 33.33\%. \citet{guo2026repetition} showed that frontier models copy random strings almost flawlessly, but on long repeats of a single character they either fail to stop or cut the answer short: on a string of 300 identical characters Qwen3.5 output 79\,984 characters in one case and 224 in another. The authors proved that a model in which the position is conveyed only by an additive bias to the attention weights that stops changing at large distances can no longer distinguish places within a sufficiently long repeat. In all these works the loss of place is not separated from other errors: the whole span, the whole string or the omitted fragment counts as an error. The synchronization introduced in this work measures the loss of correspondence of the words of the answer with their positions in the reference separately from replacement errors.

In many applied tasks the quality declines as the load grows, but the kind of the errors is not determined. \citet{kate2025longfunceval} studied tool calls and found a decline in quality by 7--85\% as the number of available tools grew and by 7--91\% as the length of their responses grew. In the work of \citet{ferguson2026extractbench} on extracting data from documents according to a given schema, with a schema of 369 fields none of the tested models produced a usable answer. \citet{xing2025mmtu} found that the accuracy of GPT-4o in information retrieval from tables dropped from 0.81 for tables of 50 rows by 5 columns to 0.43 for tables of 5 rows by 50 columns. \citet{gamage2026omission} studied the compliance of an agent with behavior rules set in the system prompt: for Mistral Large~3 the share of observed prohibitions dropped from 73\% at the fifth turn of a dialogue to 33\% at the sixteenth, while mandatory actions continued to be performed in 100\% of cases. In motion graphics generation \citep{ma2026motiongraphics} a request combines 7 to 14 kinds of requirements, and the fulfilment of each of them is checked by a program. Of the failed animations, 93.6\% pass at least one check and 31.6\% fail exactly one, and in 8.3\% of pools of 32 attempts every check passes in at least one attempt, yet no animation passes all checks at once. \citet{jiang2026depth} merged the requirements of all steps of a terminal task into one instruction, keeping the environment and the amount of work, and nine of 25 tasks were no longer solved within the original limit on the number of turns. In these works the decline in quality is established, but it is not determined whether the model lost its place in the answer, fulfilled a requirement wrongly or changed something that did not need to be changed. The work of \citet{laban2026delegate} comes closest to separating the kinds of errors: the losses of quality over a long sequence of document edits are divided into the loss and the distortion of content. About 80\% of the losses fall on rare but severe failures. However, the analysis is carried out at the level of the document, with a separate evaluation of its regions, without independently distinguishing different kinds of load.

The third kind of errors is a change of what did not need to be changed. \citet{zhu2026overediting} studied the repair of errors in program code, building the tasks so that the minimal edit is known in advance, and found that models edit more than the error requires. GPT-5.5 with a high reasoning level repaired code correctly in 82.3\% of the tasks, yet the amount of its unnecessary edits exceeded that of Opus~4.7 more than four times. A single instruction to preserve the original code reduced the mean excess edit distance from 0.195 to 0.131. The misfires considered in this work are an error of the same kind, measured on every background word. In the work of \citet{zhu2026overediting} an unnecessary edit is measured by a distance over the whole answer and is not related to the load.

The tasks of one dataset differ in difficulty, and this spread must be taken into account when estimating uncertainty. \citet{messing2026hidden} included the tasks and the prompt variants in a mixed model as random effects. On MMLU the spread of task difficulty within categories gave 72.2\% of the variance of the accuracy estimate, whereas the three prompt variables together gave 2.7\%. If these components are ignored, the 95\% confidence interval at the smallest sample sizes covers the true value in only 91--93\% of cases. The random request intercept in the statistical models of this work accounts for the same spread. The difference is that in \citet{messing2026hidden} the variance of the random effect is constant, whereas in this work it depends on the load. A model with a heterogeneous variance of the random effect is known in statistics \citep{hedeker2008application}, but among the reviewed works on the evaluation of language models it has not been applied.

Finally, models differ not only in the level of quality but also in the character of its decline. \citet{ravulapalli2026binding} measured the number of ``entity---value'' bindings that a model is able to hold in context, and for the models studied this limit ranged from 3 to 24 bindings. In IFScale \citep{jaroslawicz2025ifscale} the decline in quality with the number of instructions has a threshold, linear or exponential form depending on the model. \citet{dongre2026attention} forcibly closed the attention of a model to the instructions in a long dialogue and found that models of different architectures fail differently: Mistral partly keeps the prescribed behavior and loses it gradually as the complexity of the goal grows, whereas LLaMA and Qwen fail equally at all levels of complexity. Two more works show that the differences between models depend on the load. In Switch-SWE \citep{chang2026switchswe} switching between unfinished tasks lowers the share of solved focus tasks by 11.5 percentage points on average, and GLM-5.3 and Kimi K3, nearly equal without switching (49.0 and 48.0\%), diverge to 43.0 and 34.5\% with switching. In the transition to the harder version of AutomationBench \citep{dong2026toolspace} the quality declines for all seven models, and the gap between the strongest and the weakest narrows from about 31 to 8 points. These observations are consistent with the view that at moderate load the models diverge, and beyond the limits of each of them they converge to a common lower level. Therefore a comparison of models at a single point of difficulty is ambiguous, and models should be compared by the laws of the dependence of quality on the load. In the passports of the seven models considered in Section~\ref{sec:passports}, on part of the axes the very form of the laws differs.

In the reviewed works the load is of different nature and is set in different ways, and its influence is not separated from other factors that can cause errors. None of them separates the loss of place, the wrong execution and the misfires on the same answers, sets the kinds of load independently of each other and compares models by the laws of degradation. These tasks are addressed in this work.


\section{From One Outcome to Three Quantities}\label{sec:three-quantities}

In the previous work \citep{kutomanov2026capacity} the outcome of a trial combined two phenomena---the erroneous execution of a rule and the loss of place in the text, which manifested itself in a violation of the order of lines or a cut-off of the text. Changes of background words, to which no rule applied, remained outside the analysis. To study them, the method of matching the answer with the reference and the range of loads used in fitting the laws had to be revised.

\subsection{The former outcome of a trial}\label{sec:old-outcome}

In the previous work a trial was every occurrence in the task text of a word subject to replacement by a rule. The outcome of a trial $z$ took the value one if the correct replacement stood in the place of this word in the answer, and zero in all other cases. To determine the outcome, each line of the answer was compared with the line of the reference with the same number. With equal numbers of words in the lines, the words were compared by position. With different numbers of words only the matching initial segment was taken into account, after which the matching resumed from the next line.

The method rests on the assumption that the model rewrites the text line by line, keeping their order. As long as this assumption holds, the lines of the answer correspond to the lines of the reference with the same numbers. Deviations of the answer from the reference were described by three kinds of failure. In a failure of the third kind the answer lost its structure, and further matching became impossible. Therefore all trials after the point of failure were counted as unsuccessful \citep[Sections~3.4 and~5.4]{kutomanov2026capacity}.

Thus a zero outcome of a trial could mean either a replacement error or the impossibility of checking its correctness because of a violation of the structure of the answer. A complete loss of the original structure of the answer was treated as a collapse of the answer outside the region of measuring the instruction capacity \citep[Section~5.3.3]{kutomanov2026capacity}.

\subsection{Background words}\label{sec:background-words}

Dipankar Sarkar \citep{sarkar2026comment} independently analyzed the published results of the GPT-4o mini measurement and pointed to errors that the former way of evaluation left unobserved. These errors concerned background words---words of the source text to which no replacement rule applies. By the conditions of the task such words must remain unchanged. Since background words were not among the trials, their erroneous replacements were not counted. By construction of the task, the right words of the rules are absent from the source text \citep[Section~3.3]{kutomanov2026capacity}. The appearance of the right word of some rule in the place of a background word means that the rule was applied to a word for which it is not intended. Such an event is called a misfire.

The analysis included the answers in which the number of words in every line coincided with the number of words in the corresponding line of the reference (38\% of the answers). This made it possible to match words by position without additional alignment. The analysis showed that misfires were a widespread kind of error that the former outcome of a trial did not register. Replacements of background words occurred more often than unsuccessful trials, and in 95\% of cases they were substitutions of the right word of a rule. In 66\% of the answers with replacements of background words all trials were successful. Misfires were also observed in tasks with a single rule, where the number of replacements of background words exceeded the number of unsuccessful trials more than 20 times.

A check on the published data confirmed these results. As the author of the analysis noted, selecting the answers with a preserved line structure biased the sample towards less damaged answers. Therefore the study of misfires had to be extended to all answers of GPT-4o mini from the previous work.

\subsection{Loss of line order}\label{sec:line-order}

The analysis of all these answers showed that in the line-by-line comparison the replacements of background words are mixed with errors caused by a violation of the structure of the answer. Of the background words that the line-by-line comparison marks as damaged, 74\% are located after the point of a failure of the third kind, where a direct matching of positions is no longer possible.

The correspondence between lines can be violated even before the detected point of failure. The answers contain lines in which more than half of the background words are changed. Such lines occur almost exclusively in answers with a loss of the text. The median distance from them to the detected point of failure is about 160 lines. These lines account for 99\% of the replacements of background words by other words of the source text and for 33\% of the misfires. Hence the correspondence between the lines of the answer and the reference can be violated long before the detected point of failure and remain unnoticed when lines are compared by their numbers.

The cause was established by matching the lines of the answer and the reference by content, regardless of their numbers. The anchors are the lines that coincide with the reference exactly. Between them, omissions, insertions and replacements of whole lines are allowed. In most answers the discrepancies reduce to errors in individual words with the line structure preserved. The violations of line order are given in Table~\ref{tab:line-order}. Returns to previously written fragments and loops occur mainly in the longest texts with a small number of rules. Returns account for 76\% of the lines with mass replacements, loops for 21\%.

In 62\% of the answers at least one line changed its length, turned out to be a copy of another line, was repeated or was omitted. Hence the assumption that the line correspondence is preserved is violated in most answers, and a comparison by line numbers alone does not make it possible to evaluate replacement errors reliably.

\begin{table}[htbp]
\centering
\small
\caption{Violations of line order in the answers of GPT-4o mini from the measurement of the previous work.}\label{tab:line-order}
\begin{tabular}{@{}p{9.2cm}rr@{}}
\toprule
Violation & Discrepancies & Answers \\
\midrule
Merge of two adjacent lines or a permutation of lines & 297 & 253 \\
Cut-off (the end of the answer is missing) & 423 & 423 \\
Skip forward (the answer continues after an omitted fragment of the text) & 414 & 351 \\
Short copies of later lines & 10 & 10 \\
Return backward (the model writes a previously written fragment again) & 80 & 79 \\
Loop (one line or one word is repeated hundreds of times) & 77 & 77 \\
\bottomrule
\end{tabular}
\end{table}

In such answers the line-by-line comparison faces three problems. First, after the model returns to a previously written fragment, the lines of the answer are still matched with the lines of the reference by their numbers, although their content no longer corresponds. Such matching produces lines with mass replacements.

Second, in part of the answers the detected point of a failure of the third kind turned out to lie outside the text. In addition, after the point of failure the line-by-line comparison counted random coincidences as successful trials, although by the definition of the outcome all trials after this point had to be counted as unsuccessful.

Third, the line-by-line comparison shifts the correspondence of lines when a fragment is inserted. For example, with a reference of seven lines, an answer with the sequence of lines ``1 2 3 4 2 3 4 5 6 7'' contains all lines of the reference in the right order and additionally repeats lines 2--4. The line-by-line comparison matches the repeated lines 2--4 with lines 5--7 of the reference, counts the corresponding trials as unsuccessful and marks line 5 as the point of a failure of the third kind. Lines 5--7 of the answer itself remain unchecked. When matching by content, the first occurrences of the lines are matched with the reference, and the repeated fragment is treated as an insertion. If the replacements are performed correctly, all trials turn out successful. The violation of the structure of the answer remains outside the outcome of a trial in this case too, since the words of the inserted lines have no counterparts in the reference.

Thus the comparison of lines by their numbers does not make it possible to distinguish reliably an erroneous replacement of a background word from a violation of the correspondence between the lines of the answer and the reference. Therefore the preservation of the synchronization of the answer with the source text had to be singled out as an independent characteristic and measured separately.

\subsection{Three quantities}\label{sec:quantities}

To separate replacement errors from violations of the structure of the answer, the answer is matched with the reference by a method of sequence alignment (Section~\ref{sec:alignment}). The unit of account is a position corresponding to a word of the reference or to a word of the answer for which no counterpart was found in the reference. Three cases are distinguished in the alignment. In a match, a word of the reference corresponds to a word of the answer regardless of its correctness. In an omission, no word of the answer corresponds to a word of the reference. In an insertion, a word of the answer has no counterpart in the reference. A wrong word in a matched position counts as an error of task execution.

Three quantities are defined on the basis of the alignment.
\begin{itemize}
\item Synchronization $\Psync$ is the share of matched positions among all positions taken into account. It is computed as
\[
\Psync = \frac{n_{+} + n_{-}}{n_{+} + n_{-} + n_{\mathrm{del}} + n_{\mathrm{ins}}},
\]
where $n_{+}$ and $n_{-}$ are the numbers of correct and wrong words in matched positions, $n_{\mathrm{del}}$ is the number of omissions, and $n_{\mathrm{ins}}$ is the number of insertions. A wrong word in a matched position is counted separately from violations of synchronization. Including insertions in the denominator keeps $\Psync$ within the interval from zero to one.

\item Rule execution $H$ is the probability of a correct replacement of a word to which a rule applies, provided that the position of this word is matched with a position in the reference.

\item Misfires $F$ is the probability that the right word of some rule appears in the place of a background word, provided that the position of this word is matched with a position in the reference.
\end{itemize}

The three quantities are combined by the formula of total probability
\[
P_{\mathrm{correct}} = \Psync \bigl[\pi H + (1 - \pi)(1 - F)\bigr],
\]
where $\pi = m/D = p$ is the share of words to which the replacement rules apply, that is, the replacement density. The replacement density is set by the parameters of the request and is not estimated in fitting the model. The quantity $P_{\mathrm{correct}}$ is the expected share of correct positions of the text for a request as a whole at a given point of load.

The three quantities are distinguished because of the different nature of the corresponding errors. A violation of the correspondence of positions is counted separately as a loss of synchronization. The correctness of rule execution and the misfires are evaluated only for words matched with positions in the reference. Inserted words have no such positions, so trial outcomes are not defined for them. In addition, the failures of rule execution and the misfires depend on the load differently. In all seven models the law of the replacement density for $H$ and for $F$ differs in form (Section~\ref{sec:passports}), and in Gemini 3.5 Flash Lite and Gemini 3.1 Flash Lite the full density acts on these quantities in opposite directions: the measure of the density is below one for $H$ and above one for $F$ (Subsection~\ref{sec:passports-density}). The application of a wrong rule is counted among the failures of $H$ and makes up 14\% of these failures.

The separation of errors into three components changes the interpretation of the previous measurement substantially. No more than 16\% of the failures registered by the former way of evaluation on the GPT-4o mini data can be attributed to wrong replacements with the correspondence of positions preserved. With such correspondence a rule is violated in only 5\% of the trials of $H$. The remaining failures are related to the loss of correspondence of words with their positions or to the impossibility of a correct matching in the former line-by-line comparison.

\subsection{Collapse of the answer and the law of synchronization}\label{sec:collapse-boundary}

In the previous work the collapse of the answer was treated as an event at the level of the request as a whole. After the point of a failure of the third kind the outcome of every trial was counted as zero. The collapse was described by separate laws characterizing the probability of losing the text and the place in the answer where this happens. Regions of load in which all answers ended in a collapse were assigned to the region of the collapse of the answer. Requests from these regions were excluded from the fit of the statistical model of the instruction capacity \citep[Section~5.3.3]{kutomanov2026capacity}.

The alignment of the answer with the reference removes the need for such a division. The outcome is evaluated at every position. Lost words are counted in $\Psync$, while words that have kept the correspondence with their positions continue to take part in the evaluation of $H$ and $F$. The collapse of the answer becomes an extreme case of the loss of synchronization. The law of synchronization covers the whole range from answers with a fully preserved structure to answers that have lost it entirely, and replaces the separate laws of the collapse. Therefore the regions of load in which all requests of a control point ended in a failure of the third kind are included in the joint fit of the laws of all three quantities. These regions, defined by doublings of the parameters $D$, $n$ and $m$, are called the collapse boundary.

Including the collapse boundary substantially affects the estimate of the law of synchronization. In the GPT-4o mini measurement this boundary covers 32\% of the requests. It accounts for 80\% of the failures of synchronization and rule execution and for 54\% of the misfires. If these regions were excluded, the law would describe only a moderate degradation. The plateau of the slices would decrease by $1.19 \pm 0.52$ logits per doubling of the text length, compared with the estimate of $1.86 \pm 0.34$ logits over all requests. The measure of the density would be 17 times smaller, and the fatigue capacity would be overestimated by 29\% (Table~\ref{tab:collapse-boundary}).

\begin{table}[htbp]
\centering
\small
\caption{Laws and capacities of $\Psync$ for GPT-4o mini over all requests of the measurement and without the collapse boundary, code of release 0.2.1. The uncertainty is the half-width of the 95\% interval. The 95\% interval is given in parentheses. An asterisk marks a continuation of the law.}\label{tab:collapse-boundary}
\resizebox{\linewidth}{!}{%
\begin{tabular}{@{}lrr@{}}
\toprule
& All requests & Without the collapse boundary \\
\midrule
Requests in the fit & 5162 & 3509 \\
Mean random-intercept variance $\sigma^2$ & $12.36 \pm 0.73$ & $14.08 \pm 1.15$ \\
Law of the plateau over the text length & line & line \\
Slope of the plateau, logits per doubling of the length & $-1.86 \pm 0.34$ & $-1.19 \pm 0.52$ \\
Law of the density & binary exponential & binary exponential \\
Reference level $\alpha$ & $37.20 \pm 2.49$ & $29.60 \pm 3.36$ \\
Fatigue capacity, words & 13\,515* (10\,993--16\,617) & 17\,492* (12\,880--23\,755) \\
Measure of the density & $1.6\cdot 10^{8}$ ($6.1\cdot 10^{7}$--$4.4\cdot 10^{8}$) & $9.5\cdot 10^{6}$ ($1.6\cdot 10^{6}$--$5.7\cdot 10^{7}$) \\
\bottomrule
\end{tabular}%
}
\end{table}

Thus the law of synchronization replaces the former laws of the collapse and describes the behavior of the model up to a complete loss of the structure of the answer. Excluding the collapse boundary would leave outside the analysis the regions of load with the strongest violation of synchronization, and the laws would no longer reflect deep degradation. The laws and capacities of the seven models are given in Section~\ref{sec:passports}.

\section{Mechanics and Verification of the Alignment}\label{sec:alignment}

When the answer is aligned with the reference, every word of the reference and every word of the answer belongs to exactly one position, whatever the degree of damage of the answer. The alignment is performed in two stages. First the lines are matched, then the words within the matched lines.

\subsection{Method}\label{sec:alignment-method}
The alignment of two sequences is a known method. What is new in this work is its application to the answers of a language model.
\citet{levenshtein1966binary} introduced the measure equal to the smallest number of insertions, omissions and substitutions needed to
transform one sequence into another. \citet{needleman1970general} proposed global alignment by dynamic programming. The comparison of
biological sequences is based on these methods \citep{durbin1998biological}. In speech recognition the output of a system is aligned
with the reference text in the same way. This is how, for example, the \texttt{sclite} program from the NIST toolkit works
\citep{nist2009sctk}. In the same field \citet{morris2004wer} introduced the match error rate
\[
\mathrm{MER} = \frac{n_{-} + n_{\mathrm{del}} + n_{\mathrm{ins}}}{n_{+} + n_{-} + n_{\mathrm{del}} + n_{\mathrm{ins}}}.
\]
The denominator of MER coincides with the denominator of $\Psync$ (Subsection~\ref{sec:quantities}). In computing MER a wrong word in a
matched position is counted as an error. In computing $\Psync$ this position counts as matched, and the wrong word is counted in the
outcome of a trial of another quantity. In the evaluation of machine translation of long documents the alignment with the reference is
also used to account for omitted and extra text \citep{wang2025segale}.

\subsection{Lines and anchors}\label{sec:anchors}
Lines are compared in a canonical form, in which the right word of every rule is replaced by the corresponding left word. In the
canonical form a correct replacement and an unreplaced word coincide. Therefore a line rewritten without errors in the background words
coincides with the line of the source text regardless of how the rules were executed. Lines that coincide exactly are called anchors.
The anchors form the longest common subsequence of the lines of the answer and the reference. Among equivalent variants the one is
chosen in which the lines of the reference correspond to earlier lines of the answer. Therefore, on a return backward, the lines of the
first pass are matched with the reference, and the repeated fragment counts as an insertion.

\subsection{Gaps between anchors}\label{sec:gaps}
The lines of the reference and the answer between adjacent anchors form a gap. A line of the answer that exactly repeats a line of the
reference counts as an insertion in this gap, since the repeated line of the reference is already matched. The remaining lines of the gap
are matched so that the number of coinciding words is the largest. Three kinds of matching are allowed. One line of the answer may
correspond to one line of the reference or to two adjacent lines of the reference (a merge). Two lines of the answer may correspond to one
line of the reference (a split).

A matching is allowed only if more than half of the words of the shorter side coincide:
\[
2c > \min(a, r),
\]
where $c$ is the number of coinciding words, and $a$ and $r$ are the numbers of words on the side of the answer and of the reference,
respectively. This condition is called the copy threshold. The probability that a word of another line coincides with a word of the
reference by chance is about $1/D$. The words of a rewritten line almost always coincide with the words of the reference. The value of a
matching is $2c$. For a merge and a split it is reduced by three, so these variants are chosen only if they give at least two more
coinciding words than matching one line with one. A line of the answer left without a pair counts as an insertion. A line of the
reference without a pair counts as an omission.

\subsection{Words, wrapper and word number}\label{sec:words}
In matched lines of equal length the words are compared by position, as in the line-by-line comparison. In the other pairs the
alignment with the smallest number of edits is chosen. A substitution, an omission and an insertion count as one edit each.

Lines at the beginning and at the end of the answer that contain no word of the task form a wrapper and are excluded from the account.
The wrapper includes, for example, introductory phrases of the model and the delimiters of a code block. The word number of every
position is counted by the reference. An insertion is assigned the number of the word that follows, in the reference, the last
coinciding word before the insertion in the answer.

\subsection{Contribution of failures to synchronization}\label{sec:failure-cost}
The contribution of every failure is determined by the number of positions it affects. The omission of one word gives one unmatched
position, and a permutation of two adjacent lines of ten words each gives twenty. In a merge the number of unmatched positions equals the
number of lost words. A split without a loss of words creates no unmatched positions. With an omission of lines or a cut-off, every word
of the omitted lines gives one unmatched position. With a repeat or a return backward, every inserted word gives one such position. Thus
$\Psync$ measures the loss of place in the text by the number of affected words. For every answer it is checked that every word of the
reference and every word of the answer belongs to exactly one position. A violation of this condition would indicate an error of the
instrument.

\subsection{Verification on the measurement of the previous work}\label{sec:alignment-check}
The alignment is applied to all answers of the GPT-4o mini measurement from the previous work. The results are given in
Table~\ref{tab:places}. The sum of the number of matched positions and the number of omissions equals exactly the number of words in all
texts of the measurement.

\begin{table}[htbp]
\centering
\small
\caption{Positions and trials in the GPT-4o mini measurement from the previous work.}\label{tab:places}
\begin{tabular}{@{}p{8.4cm}r@{}}
\toprule
Quantity & Value \\
\midrule
All positions & 8\,186\,622 \\
Matched positions & 5\,470\,049 \\
Omissions & 1\,077\,371 \\
Insertions & 1\,639\,202 \\
Share of matched positions & 0.668 \\
Trials of $H$ & 531\,861 \\
Share of correct replacements among them & 0.946 \\
Trials of $F$ & 4\,938\,188 \\
Share of misfires among them & 0.0066 \\
\bottomrule
\end{tabular}
\end{table}

For the answers in which the line-by-line comparison found no violations of structure, the assumption of the former method holds. In
these cases the alignment must reproduce the result of the line-by-line comparison. At every matched position both methods must give the
same word and the same trial outcome. The check covered all such answers (38\% of the answers of the measurement). In all trials both the
words and the outcomes coincided. The implementation of the former method on the same measurement exactly reproduced the published
numbers of trials and failures.

Gaps with omissions or insertions are found in 62\% of the answers. This agrees with the share of answers with violations of line order
obtained in Subsection~\ref{sec:line-order} by matching whole lines.

\subsection{Limitations of the method}\label{sec:alignment-limits}
With a small number of distinct words $D$, random coincidences can pass the copy threshold. In the GPT-4o mini measurement from the
previous work the share of coinciding words was between 0.5 and 0.7 for 2.4\% of the line matchings that passed the threshold and for
45\% of the merges. Thus merges are the most exposed to this error. Whether a rigid threshold can be replaced by a probabilistic
evaluation of doubtful matchings remains an open question. Merges and splits are considered only for two lines. A line with many wrong
words does not pass the threshold and counts as an omission and an insertion. A permutation of two adjacent words in a line of the same
length counts as two wrong words in matched positions rather than as a violation of synchronization. A repeat and a return backward are
counted in the same way, as an insertion.

For an insertion the word number is determined by the reference. This choice has not been checked. The law of fatigue may require the word
number in the answer. The alignment takes into account the order of words. Conclusions about the mechanism of the errors of the model are
limited to what can be established from the matching of positions.

\section{Statistical Models}\label{sec:models}

The statistical apparatus of the previous work \citep{kutomanov2026capacity} is retained. What changes is the model of the
random-intercept variance: the variance now depends on the load (Section~\ref{sec:spread}). The previous work modeled one trial outcome.
This work builds three statistical models, each with its own trials and load axes.

\subsection{Trials}\label{sec:trials}
The trials of each of the three quantities are given in Table~\ref{tab:trials}. Rule words and background words whose positions are not
matched with the reference count as failures of $\Psync$ and are excluded from the trials of $H$ and $F$. A replacement of a background
word counts as a misfire only when the right word of some rule is substituted.

\begin{table}[htbp]
\centering
\small
\caption{Trials of the three quantities.}\label{tab:trials}
\begin{tabular}{@{}p{1.6cm}p{4.2cm}p{3.6cm}p{5.0cm}@{}}
\toprule
Quantity & Trial & Success & Failure \\
\midrule
$\Psync$ & Position & Matched position & Omission or insertion \\
$H$ & Rule word in a matched position & Correct replacement & The word is left unreplaced, replaced by the right word of another rule or by some other word \\
$F$ & Background word in a matched position & No misfire & The right word of a rule in the place of the background word \\
\bottomrule
\end{tabular}
\end{table}

In every model the logit corresponds to the probability of success. For $F$ this is the logit of the probability $1 - F$. Therefore in
all three models a smaller coefficient corresponds to a larger probability of failure. The definitions of the logit, the request
intercept, the bins, the smooth laws, the shift correction and the capacities are given in Section~4 of the previous work
\citep{kutomanov2026capacity}.

\subsection{Three models}\label{sec:three-models}
The logit of the probability of success in trial $i$ of request $j$ is given by
\[
\begin{aligned}
\Psync:\quad \eta_{ij} &= VS(v, s) + P(p) + M(m) + b_j,\\
H:\quad \eta_{ij} &= VS(v, s) + P(p) + M(m) + R(r) + b_j,\\
F:\quad \eta_{ij} &= V(v) + q(s/v) + P(p) + M(m) + b_j,\\
b_j &\sim \mathcal{N}(0, \sigma_j^2).
\end{aligned}
\]
Here $v$ is the text length, $s$ is the word number, $p$ is the replacement density, $m$ is the number of rules, $r$ is the rule number,
and $s/v$ is the share of the answer. $VS$ denotes the free coefficient of a cell of the grid corresponding to a pair of bins of the text
length and the word number. $P$, $M$, $R$, $V$ and $q$ denote the free coefficients of the bins of the corresponding axes. All axes are
defined for any prompt containing instructions and an input text. This is necessary for the method to be applicable outside the
instrument \citep[Section~4]{kutomanov2026capacity}. The variance $\sigma_j^2$ of the intercept of request $j$ depends on the load of
this request. Its model is described in Section~\ref{sec:spread}.

\subsection{Choice of axes}\label{sec:axes-choice}
The axis of the number of rules is included in the model of synchronization so that its influence is measured rather than assumed
absent. On the points of the previous work, obtained on GPT-4o mini, no influence of this axis was found: the joint Wald test of the
coefficients of its bins gave $\chi^2 = 6.48$ with 8 degrees of freedom ($p$-value 0.59). This result agrees with the law of the collapse
rate of the previous work, according to which, at a given text length and replacement density, the number of rules does not affect the
probability of losing the text \citep[Section~5.4.1]{kutomanov2026capacity}.

In the measurements of the seven models the influence of the number of rules on synchronization differs. In GPT-4o mini and Gemini 3.1
Flash Lite the slope of the law of the number of rules is indistinguishable from zero. In Gemini 3.5 Flash Lite, Nemotron 3.5 Lightning
and Nemotron 3 Ultra synchronization improves as the number of rules grows, and in Nemotron 3 Nano Omni and GPT-4.1 mini it worsens, in
GPT-4.1 mini by about 0.5 logits per doubling of the number of rules. However, in these two models too the reference level of
synchronization is so high that the instruction capacity of $\Psync$ exceeds $10^{30}$ rules (Subsection~\ref{sec:laws-sync}). Thus the
number of rules may influence synchronization, but in practice does not limit it. The axis of the rule number is not part of the model of
synchronization, since a position as the unit of account of synchronization has no rule number.

The model of rule execution keeps the axes of the outcome model of the previous work. Its trials are only the rule words in matched
positions. Therefore the laws of this model describe the replacement with the correspondence of positions preserved, and their form
changes (Section~\ref{sec:passports}).

The axis of the rule number is not part of the model of misfires, since a background word has no rule number. On the points of the
previous work, obtained on GPT-4o mini, the slices over the text length hardly depend on the word number: the coefficient of a cell of the
grid deviates from the sum of the coefficients of the corresponding bins of the text length and the word number by 0.16 logits on
average. Two cells out of 49 fall outside the 95\% interval, with 2--3 expected. Introducing the grid does not change the estimates on the
other axes. Therefore the text length and the progress through the answer are described by separate axes. The progress is given by the
share of the answer.

\subsection{From models to the passport}\label{sec:models-to-passport}
The transition from the models to the passport remains the same. The fit yields profiles of the coefficients of the bins, through which
smooth laws are drawn (Section~\ref{sec:smooth-laws}). The value of a law at the zero point of the axis serves as the shift of this
axis. The sum of the shifts gives the reference level of the quantity. The capacities are derived from the laws and the reference level
\citep[Sections~4.4 and~4.5]{kutomanov2026capacity}.

The formula of total probability combines the three quantities, whose laws and capacities are determined separately
(Subsection~\ref{sec:quantities}). The laws of the collapse rate and of the place of the collapse from the previous work are excluded from
the new passport, since the loss of the structure of the answer is now described by $\Psync$ (Subsection~\ref{sec:collapse-boundary}).
The laws and capacities of the seven models are given in Section~\ref{sec:passports}.

\section{Heterogeneous Random-Intercept Variance}\label{sec:spread}

In the models of Section~\ref{sec:models} the differences between requests are described by an intercept $b_j$ common to all trials of
request $j$. In the previous work \citep{kutomanov2026capacity} the variance of this intercept was constant over the whole measurement.
This section shows that the spread between requests depends on the load, introduces a model of this dependence and describes its
influence on the laws. It also justifies the choice of the level 0.5 for computing capacities. The numerical results and figures refer
to the GPT-4o mini measurement, and the random-intercept variance of the other models is given in their passports.

\subsection{One variance does not agree with the data}\label{sec:spread-test}
When the dependence of the random-intercept variance on the load is taken into account, the gain in the log-likelihood substantially
exceeds what the added parameters can give. With a constant variance, twice the gain in the log-likelihood $2\Delta\ln L$ would follow a
$\chi^2$ distribution with the number of degrees of freedom equal to the number of added parameters. In the GPT-4o mini measurement the
observed $2\Delta\ln L$ exceeds the corresponding critical value at the significance level 0.001 by a factor of 12--27 for all three
quantities. For each of the 13 quantities fitted with a heterogeneous variance in the passports of the seven models, the observed value
exceeds the corresponding critical value at least threefold.

If the distribution of the random intercept depends on covariates while the model treats its variance as constant, the estimates of the
coefficients are biased \citep{heagerty2001misspecified}. This bias is transferred to the laws built from the coefficients of the bins of
a model with one variance.

\subsection{Model}\label{sec:spread-model}
The expression for the logit of each quantity is retained. What changes is the distribution of the intercept. Its variance is given by
the submodel
\[
b_j \sim \mathcal{N}(0, \sigma_j^2), \qquad \ln\sigma_j = z_j\gamma,
\]
where $z_j$ is a row containing one for the free term and the indicators of the groups of the text length, the density and the number of
rules of request $j$, and $\gamma$ is a vector of parameters. Such a model is called below a GLMM with heterogeneous random-intercept
variance. For the variance of the random intercept the same log-linear submodel is used as for the between-subject variance in the model
of \citet{hedeker2008application}. With a binary trial outcome there is no separate submodel of the spread within a request. The
variance of the intercept depends only on the three load axes that are constant within a request: the text length, the density and the
number of rules. The word number and the rule number change within a request and do not enter the submodel of the intercept variance.

A group of an axis consists of one bin or of several merged rare bins at the edge of the axis. Bins are merged until the group contains 30
requests with failures. The group with the largest number of requests serves as the reference. The coefficients of the groups, like the
coefficients of the bins of the logit, are estimated freely, so the form of the dependence of the variance on the load is not set in
advance. The parameters of the logit and $\gamma$ are estimated by a joint maximization of one marginal likelihood. The same fitting
procedure as for the model with one variance is used, and its estimates serve as initial values. The model with one variance is used if
the fit with a heterogeneous variance does not converge or if the resulting law contradicts the data by the criterion described in
Subsection~\ref{sec:spread-check}.

\subsection{Spread along the load axes}\label{sec:spread-axes}
\begin{figure}[htbp]
\centering
\includegraphics[width=\linewidth]{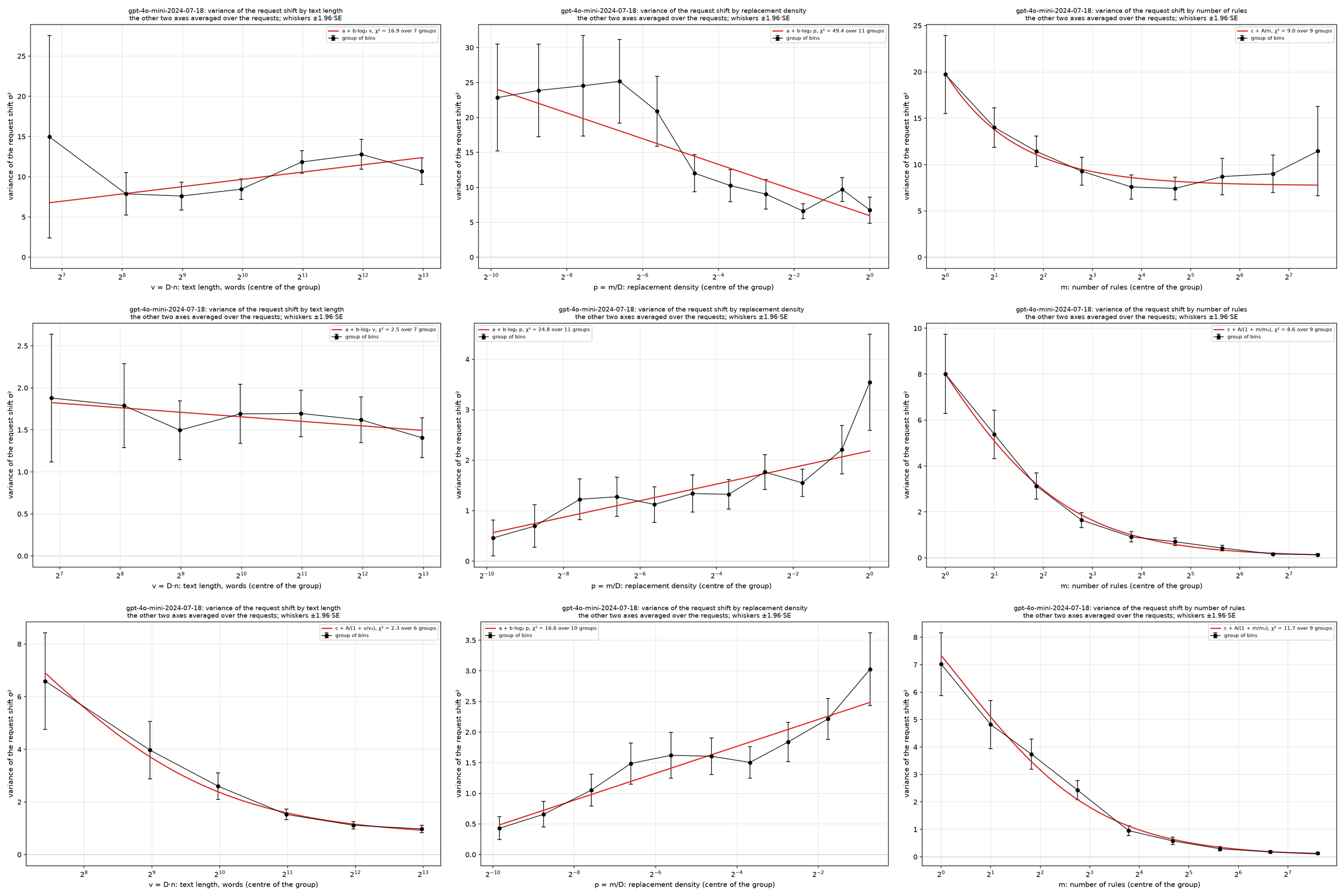}
\caption{Random-intercept variance by the groups of the axes for GPT-4o mini. The rows correspond to $\Psync$, $H$ and $F$, the columns to
the text length, the density and the number of rules. Over the two remaining axes the variance is averaged over the requests. Whiskers
show $\pm 1.96$ standard errors, curves are the approximations along the axes. For $\Psync$ the model rejected by the median-request
check is shown (Subsection~\ref{sec:spread-check}).}\label{fig:spread-grid}
\end{figure}

Figure~\ref{fig:spread-grid} shows the random-intercept variance by the groups of the three load axes. The character of the dependence
differs between the axes and the quantities. In the GPT-4o mini measurement, as the number of rules grows the spread of $H$ and $F$
decreases by factors of 62 and 54 respectively, and the spread of $\Psync$ by a factor of two to three. As the density grows the spread of
$\Psync$ decreases almost fourfold, whereas the spread of $H$ and $F$ increases 7--8 times. As the text length grows the spread of $F$
decreases almost sevenfold. The spread of $\Psync$ first decreases twofold and then grows again on long texts. The spread of $H$ hardly
changes.

The curves in the figure approximate the estimates by groups. A line over the logarithm of the axis $a + b\log_2 x$, a fractional
hyperbola $c + A/(1 + x/x_0)$ and its limiting case $c + A/x$ are used. The form of the approximation was chosen separately for each axis
of each quantity.

\subsection{Laws with a heterogeneous variance}\label{sec:spread-laws}
\begin{figure}[htbp]
\centering
\includegraphics[width=\linewidth]{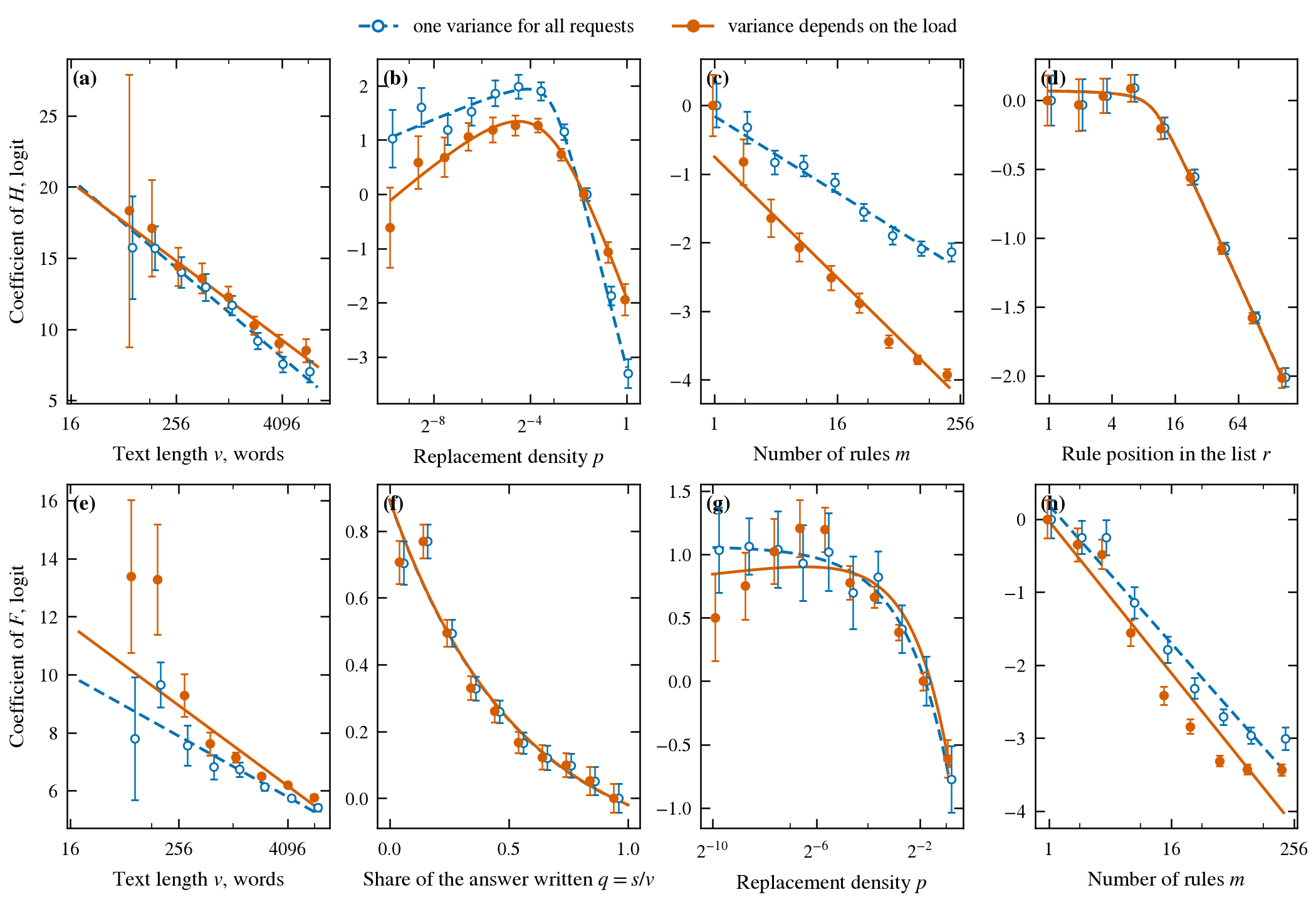}
\caption{Laws of the axes of $H$ (top row) and $F$ (bottom row) for GPT-4o mini with a constant (blue) and a heterogeneous (orange)
random-intercept variance. Points show the coefficients of the bins relative to the reference bin with 95\% intervals, lines are the
laws.}\label{fig:laws-before-after}
\end{figure}

Figure~\ref{fig:laws-before-after} shows the laws of the axes of $H$ and $F$ obtained in the fit with a constant and with a
heterogeneous variance. For $\Psync$ one variance is kept by the median-request check (Subsection~\ref{sec:spread-check}), so its laws
coincide in the two variants. The laws of the axes that change within a request also coincided: those of the rule number for $H$ and of
the share of the answer for $F$.

The laws of the load axes of the request changed, and the direction of the change is related to the dependence of the spread on the
load. Where the spread decreases as the load grows, the law with a heterogeneous variance turns out steeper. Over the whole range of the
number of rules the law of $H$ decreases by 3.4 logits with a heterogeneous variance and by 2.1 logits with a constant one. Where the
spread grows, the law turns out flatter. From the rarest replacements to the full density the law of $H$ changes by 1.7 logits with a
heterogeneous variance and by 4.3 logits with a constant one. For $F$ the laws change in the same directions.

At the logit $\eta$ of the median request, the share of successes over all requests is approximately equal to the value of the logistic
function at $\eta/\sqrt{1 + 3\sigma^2/\pi^2}$. If the variance in the model is constant, its dependence on the load is absorbed by the
estimates of the coefficients of the bins. The bias affects the coefficients of the request features along which the variance changes
\citep{heagerty2001misspecified}. The coefficients of the features that change within a cluster are only weakly affected by a
misspecified distribution of the random effects \citep{mcculloch2011misspecifying}.

\subsection{Median-request check}\label{sec:spread-check}
In the GLMM the coefficient of a bin is estimated for the median request, whose intercept is $b = 0$. The share of successes over all
requests is approximately determined by the logistic function of $\eta/\sqrt{1 + 3\sigma^2/\pi^2}$. If the random-intercept variance
changes strongly along an axis, the relation between the logit of the median request and the share of successes differs between groups.
As a result, the coefficient of a bin relative to the reference may have the sign opposite to the sign of the difference observed over
the requests. Therefore, for each quantity the estimates of the bins are checked against the median requests in the data.

The axes that are constant within a request are checked: the density, the number of rules and, for $F$, the text length. Requests
containing at least 50 trials of the given quantity are distributed over the cells of the two remaining axes. In each cell the median
shares of failures over the requests of the compared and of the reference bin are converted into logits of the probabilities of success.
The difference of the logits is averaged over the cells with weights equal to the number of requests of the reference bin in each cell.
The standard error of the difference is estimated from 200 bootstrap resamples of the requests. If the coefficient of a bin in the
passport and the difference for the median request have opposite signs, and the absolute value of the difference exceeds 1.96 of its
standard error, the law with a heterogeneous variance is considered to contradict the data. In this case the quantity is fitted with one
variance.

In the GPT-4o mini measurement the check rejected the heterogeneous variance for $\Psync$. In that fit the coefficients of five bins of
the number of rules, from 5--8 to 129--256 rules, lie below the reference and range from $-0.15$ to $-0.44$. The differences for the
median requests in the data are positive and range from $+0.73 \pm 0.52$ to $+2.81 \pm 1.01$. Therefore in the GPT-4o mini passport
$\Psync$ is fitted with one variance. The numbers for $\Psync$ in Subsections~\ref{sec:spread-test}, \ref{sec:spread-axes}
and~\ref{sec:spread-level} refer to the rejected model. For $H$ and $F$ the check kept the heterogeneous variance. Among the seven models
of Section~\ref{sec:passports}, 8 quantities out of 21 are fitted with one variance.

\subsection{Level of the capacities}\label{sec:spread-level}
The laws describe the median request with intercept $b = 0$. The capacity is defined over all requests. At the logit $\eta$ of the median
request the share of successes over all requests equals
\[
\bar\pi(\eta, \sigma^2) = E_b\Bigl[\frac{1}{1 + e^{-(\eta + b)}}\Bigr], \qquad b \sim \mathcal{N}(0, \sigma^2).
\]
The threshold of the level $q$ is the root of the equation $\bar\pi(\eta^*, \sigma^2) = q$. The reserve of the level equals the
difference between the reference level and the threshold. With a constant variance each quantity had one threshold for the whole
measurement. With a heterogeneous variance the threshold and the reserve depend on the load:
\[
\Delta_q(v, p, m) = \alpha - \eta^*\bigl(q, \sigma^2(v, p, m)\bigr).
\]
The threshold is computed numerically. Its approximate value at a given variance can be obtained by the formula
\[
\eta^*(q, \sigma^2) \approx \ln\frac{q}{1 - q}\,\sqrt{1 + \frac{3\sigma^2}{\pi^2}}.
\]
At $q = 0.95$ the right-hand side equals $2.944\sqrt{1 + 0.304\,\sigma^2}$. The formula is exact for an intercept with the bridge
distribution \citep{wang2003matching}. With a normally distributed intercept and the level 0.95 its error does not exceed 1.4\% in the
range of $\sigma^2$ from 0.5 to 50. For a level above 0.5, a larger spread at a given load corresponds to a higher threshold and a smaller
reserve. In the GPT-4o mini measurement the threshold of the level 0.95 over the groups of the axes ranges from 3.0 to 5.5 for $H$ and
$F$. With zero spread it would equal 2.94.

Therefore a capacity at a level above 0.5 depends on the load at which the variance is computed. The values of the axes that enter the
definition of a capacity are set by the corresponding load. The values of the remaining axes have to be set in addition, and this choice
strongly affects the capacity estimate. In the version of the passport with capacities at the level 0.95, the guaranteed fatigue
capacity of $F$ for GPT-4o mini, that is, the fatigue capacity at full density, was 5071 words if the number of rules was averaged over
the requests in computing the variance, and 1.9 words for one rule.

At the level 0.5 the threshold does not depend on the variance. The sum $1/(1 + e^{-b}) + 1/(1 + e^{b})$ equals one. Therefore, for a
symmetric distribution of the intercept, $\bar\pi(0, \sigma^2) = 1/2$ for any variance. The threshold of the level 0.5 is zero, and the
reserve equals the reference level $\alpha$. A capacity at the level 0.5 is determined by the load at which the logit of the median request
reaches zero. For given laws this value depends neither on the variance, nor on the choice of the point of load at which it is computed,
nor on the shape of the tails of the distribution of the intercept. Therefore all capacities in this work are given at the level 0.5. The
same principle is used in defining the half-effect dose in dose--response relationships and the difficulty of an item in test theory.

In the previous work capacities were also given at the level 0.95. The level 0.5 is far from working quality, and part of the capacities at
this level lie outside the measured region and are obtained by continuing the laws. The heterogeneous variance is taken into account in
the fit, since with a constant variance the estimates of the laws themselves are biased (Subsection~\ref{sec:spread-laws}). In computing
capacities at the level 0.5 the threshold correction is zero. The capacities of the three quantities for the seven models are given in
Section~\ref{sec:passports}.

\subsection{Why the spread changes}\label{sec:spread-why}
As the number of rules grows, the spread of $H$ decreases in all models in which this quantity is fitted with a heterogeneous variance;
the spread of $F$ also decreases on the whole, but in Gemini 3.5 Flash Lite and Nemotron 3 Ultra non-monotonically. This decline is
explained by the averaging of the difficulties of the rules. Rules differ in difficulty for the model, and the intercept of a request is
made up of the mean difficulty of its rules. The mean of several independent difficulties varies from request to request less than the
difficulty of an individual rule. Therefore, as the number of rules grows, the spread decreases and approaches a plateau corresponding to
the component of difficulty common to the whole request. For $H$ and $F$ of GPT-4o mini the initial part of the decline is flatter than
simple averaging implies, but the direction of the decline and the approach to the plateau are preserved.

For synchronization no common dependence on the number of rules is observed. In Nemotron 3.5 Lightning the spread of $\Psync$ decreases
as the number of rules grows, in Gemini 3.5 Flash Lite it changes non-monotonically, and in Gemini 3.1 Flash Lite it does not depend on
the number of rules.

The changes of the spread along the text length and the density are not explained by statistics. As the length and the density grow, the
number of trials in a request increases. In the fit the noise of the trials is separated from the spread of the intercepts, so a larger
number of trials increases the precision of the estimate of $\sigma^2$ without a systematic change of its value. Hence the growth of the
spread of $H$ and $F$ with the density and the decline of the spread of $F$ with the text length characterize the behavior of the model
itself. The load changes the degree to which the outcome depends on the particular request.

\section{Construction of Smoothed Laws}\label{sec:smooth-laws}

For each axis the fit of the statistical model estimates the coefficients of the bins and their uncertainties. The law of an axis is a
smooth curve fitted to these coefficients. From the law the passport determines the form of the dependence, the level of the block, that
is, the value of the law at the zero point of the axis, and the capacities. Constructing a law requires determining the set of bins,
their weights and the form of the curve.

In release 0.2.0 a law was fitted by generalized least squares with the full covariance matrix of the bins. A fallback form replaced the
established one only if the criterion decreased by at least 10 units. The sign of part of the parameters was restricted. On the seven
models considered in Section~\ref{sec:passports}, this approach led to systematic discrepancies between the laws and the data. The slope
of a line deviated from the trend seen in most bins. The restriction of the sign of the parameters did not make it possible to describe
increasing segments, and when a form with a pole at the zero point of the axis was used, the level of the block was determined by the
first bin.

The principles of constructing laws adopted starting from release 0.2.1 are set out below. Two of them are introduced for the first time
in this work: the clipping of outliers at three standard deviations and the weighting with weights $1/(\sigma_i^2 + \tau^2)$.

\subsection{Outliers}\label{sec:smooth-outliers}
Every bin for which an estimate of the coefficient is obtained is included in the fit of the law. Only bins classified as outliers are
excluded. Outliers are clipped by the iterative procedure of the astropy library \citep{astropy2022}. At each iteration the median and
the standard deviation are computed from the residuals of the bins not yet excluded. Bins whose residuals differ from the median by more
than three standard deviations are excluded. The law is fitted again to the remaining bins. The procedure continues until the number of
excluded bins stops changing. The clipping scale is the same for all bins and does not depend on the uncertainty of an individual bin.

\subsection{Weights}\label{sec:smooth-weights}
Every point of a law---a bin of an axis, a cell of the grid or a slice level---enters the fit with the weight
\[
w_i = \frac{1}{\sigma_i^2 + \tau^2},
\]
where $\sigma_i^2$ is the own variance of the point, and $\tau^2$ is the spread of the points around the law beyond their own variances.
The quantity $\tau^2$ is estimated by the method of DerSimonian and Laird in its meta-regression version
\citep{dersimonian1986meta,viechtbauer2010outlier}:
\[
\tau^2 = \max\Bigl(0,\ \frac{\chi^2 - (k - p)}{\operatorname{tr} P}\Bigr), \qquad P = W - WJ\,(J^{\top}WJ)^{+}J^{\top}W, \qquad
W = \operatorname{diag}(1/\sigma_i^2),
\]
where $J$ is the matrix of the derivatives of the law with respect to its free parameters, $k$ is the number of points, $p$ is the rank of
$J$, and $\chi^2$ is the weighted sum of squared residuals with the weight matrix $W$. The quantity $\tau^2$ is estimated for the candidate
form with the smallest value of $\chi^2$ per degree of freedom. This avoids counting systematic deviations caused by the curvature of the
dependence as spread of the points. All candidates of a given block are fitted with the same weights.

The spread of the bins around any smooth law exceeds their own uncertainties. With weights $1/\sigma_i^2$, and especially with the full
covariance matrix, the law is determined by a few most precise bins. With an excess spread it systematically deviates from most of the
bins. The addition of $\tau^2$ equalizes the contributions of the points to the extent that the spread exceeds their own uncertainties.
If the spread is within the uncertainties, $\tau^2 = 0$ and the weights are kept. For example, for GPT-4.1 mini the slope of the law of
the number of rules for $H$ with the weights $1/(\sigma_i^2 + \tau^2)$ is $-0.56 \pm 0.06$ logits per doubling of the number of rules. The
slope of the former law was $-0.27$ logits per doubling and did not agree with the coefficients of the bins.

The weights are computed without the correlations between the points, since with the correlations the law of a slice deviates from the
cells of this slice. The uncertainties of the laws are, as before, computed from the full covariance matrix of the coefficients through
the sensitivity of the estimates of the parameters of the law to these coefficients.

\subsection{Form of a law}\label{sec:smooth-forms}
For each block an established form and fallback forms are set. The fallback forms include a line, a slanted hyperbola and a binary
exponential. For the density of $H$ the two-slope hyperbola is established, for the density of $F$ a logarithm with a line; the three
named forms serve as their fallbacks. For the share of the answer of $F$ an exponential is established, and its only fallback form is a
line. For the slices of the grid the form is chosen from the two-asymptote hyperbola, the slanted hyperbola and the binary exponential.
The parameters of the curve are common to all slices, and the level is estimated for each slice separately.

All forms are fitted to the remaining bins with the same weights. The form with the smallest value of the criterion
\citep{burnham2002model}
\[
\mathrm{QAIC} = \chi^2/\hat c + 2k
\]
is chosen, where $k$ is the number of free parameters and $\hat c$ is the overdispersion coefficient. The coefficient $\hat c$ equals the
smallest value of $\chi^2$ per degree of freedom among the forms for which the fit converged, but is taken to be no smaller than one. Using
the smallest value prevents $\hat c$ from being inflated by a poorly fitted form. With equal values of the criterion the established form
is chosen.

Three rules apply to the choice of the form. First, the amplitudes of the exponential and the binary exponential, as well as the slope of
the slanted hyperbola, may have any sign. This makes it possible to describe both the decline and the growth of quality as the load
increases. Increasing segments are observed in the data (Subsection~\ref{sec:passports-rises}).

Second, a fallback form one of whose parameters reaches the boundary of the admissible range does not replace the established one. In
this case the estimate corresponds to the boundary of the family, and the value of the criterion says little about the suitability of the
form itself.

Third, the simple hyperbola $c + A/x$ is excluded from the forms considered. The level of each block is determined by the value of the law
at the zero point of the corresponding axis. For the different axes this is one rule, the first rule of the list, a text length of 20
words, the density $2^{-10}$ or the beginning of the answer. The simple hyperbola has a pole at the zero point of the abscissa, so the
level of the block would be determined by the first bin. For the same reason a saturation curve is excluded from the forms. In the
coordinates of the logarithm of the load its steepness is largest near the zero point, and its left segment is determined by the data of
the right part of the axis.

For every law the passport gives the chosen and the established forms, the value of $\hat c$, the results of fitting every candidate and
the list of excluded bins. From these data the choice of the form can be checked.

\section{Passports of Seven Models}\label{sec:passports}

Following the protocol set out in the previous sections, seven models from three developers were measured. OpenAI is represented by
GPT-4o mini and GPT-4.1 mini, Google by Gemini 3.5 Flash Lite and Gemini 3.1 Flash Lite, NVIDIA by Nemotron 3.5 Lightning, Nemotron 3
Nano Omni and Nemotron 3 Ultra. The number of measured points per model ranges from 3695 to 6961.

All measurements used the same tasks and one method of aligning the answers with the reference. The passports are built by the same code
following the principles set out in Sections~\ref{sec:spread} and~\ref{sec:smooth-laws}. The common procedure makes the passports of the
models comparable.

{\sloppy The passports contain the laws along all axes, the coefficients of the bins and the figures. They are published in the open
repository \url{https://github.com/kutomanovhennadii/instruction-capacity} (release 0.2.3) and in a collection of datasets on Hugging Face
\url{https://huggingface.co/collections/kutomanovhennadii/iec-passports-6ae3ad9fe2c798814eb8a3dc}. The measured points are published for
four models. The passports of the three Nemotron models are published without the measured points (Section~\ref{sec:availability}).\par}

The laws of the seven models are brought together on common charts, and the capacities and the measures of the density in one table. The
uncertainty is given as the half-width of the 95\% confidence interval. The interval itself is given in parentheses. An asterisk marks
estimates obtained by continuing a law beyond the measured region. The capacities are computed at the level 0.5
(Subsection~\ref{sec:spread-level}).

\subsection{Summary charts of the laws}\label{sec:laws-forms}
Every chart except the charts of the word position in the answer shows the coefficients of the bins with 95\% intervals and the laws of
the seven models for one axis of one quantity. After the shift correction the value of the law of every model at the zero point of the
axis is zero. Bins with intervals wider than 100 logits are omitted, since the fit does not determine their level.

The dependence on the word position in the answer is described by the slices of the grid. The parameters of the curve are common to all
slices, and every slice has its own level. The summary chart shows the form of the law of a slice, that is, the change of the logit
relative to the beginning of the answer. The forms are chosen from one set of curves by the smallest value of the QAIC criterion
(Section~\ref{sec:smooth-laws}). On some axes the chosen forms differ between the models.

\begin{figure}[htbp]
\centering
\includegraphics[width=\linewidth]{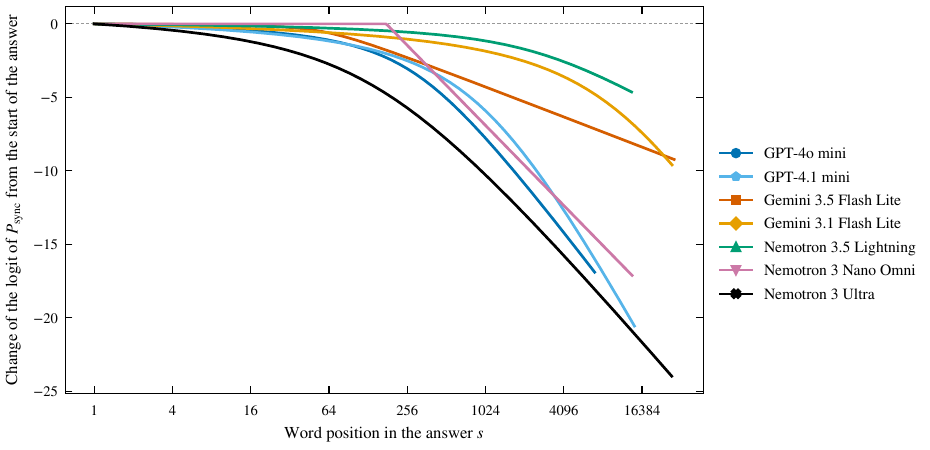}
\caption{Synchronization $\Psync$ of seven models. Form of the law of a slice over the word position in the answer. The change of the
logit relative to the beginning of the answer is shown.}\label{fig:psync-position}
\end{figure}

\begin{figure}[htbp]
\centering
\includegraphics[width=\linewidth]{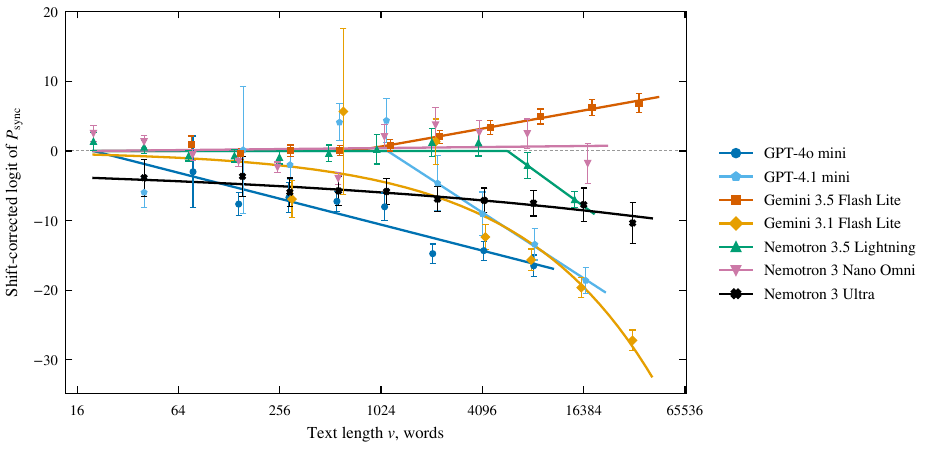}
\caption{Synchronization $\Psync$ of seven models. Slice levels over the text length and their laws.}\label{fig:psync-length}
\end{figure}

\begin{figure}[htbp]
\centering
\includegraphics[width=\linewidth]{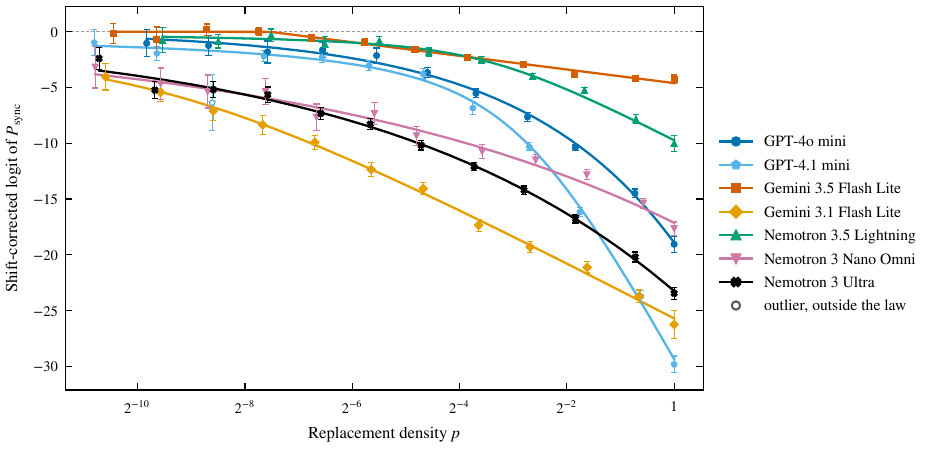}
\caption{Synchronization $\Psync$ of seven models. Laws of the replacement density.}\label{fig:psync-density}
\end{figure}

\begin{figure}[htbp]
\centering
\includegraphics[width=\linewidth]{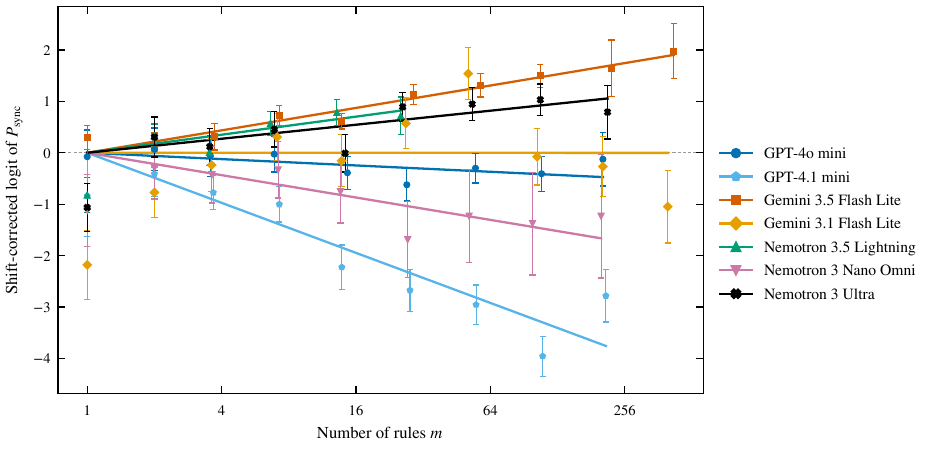}
\caption{Synchronization $\Psync$ of seven models. Laws of the number of rules.}\label{fig:psync-rules}
\end{figure}

\paragraph{Synchronization.}\label{sec:laws-sync}
The laws of synchronization are shown in Figures~\ref{fig:psync-position}--\ref{fig:psync-rules}. In all models synchronization worsens
along the answer and as the replacement density grows. As the text length increases, the slice levels decrease in five models, hardly
change in Nemotron 3 Nano Omni and grow in Gemini 3.5 Flash Lite.

For the axis of the number of rules the QAIC criterion chose, in all models, a line over the logarithm of the number of rules out of three
forms. In GPT-4o mini and Gemini 3.1 Flash Lite its slope is indistinguishable from zero. As the number of rules grows, synchronization
improves in Gemini 3.5 Flash Lite, Nemotron 3.5 Lightning and Nemotron 3 Ultra and worsens in Nemotron 3 Nano Omni and GPT-4.1 mini.

\begin{figure}[htbp]
\centering
\includegraphics[width=\linewidth]{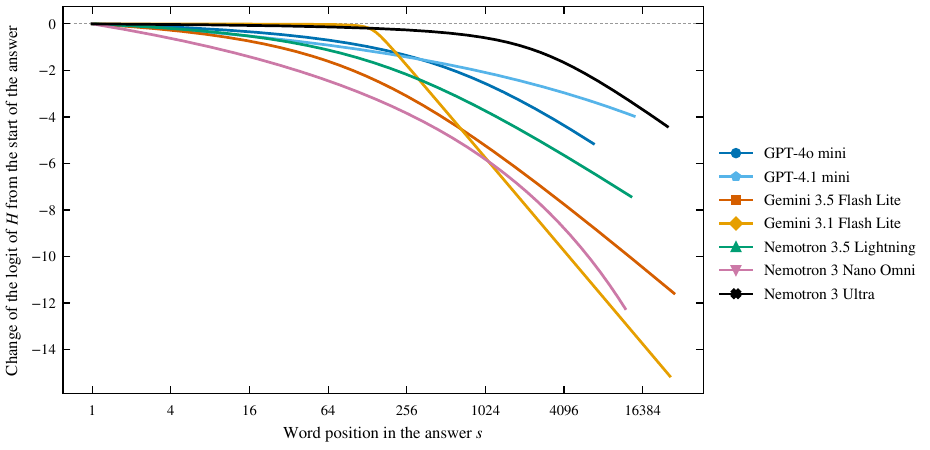}
\caption{Rule execution $H$ of seven models. Form of the law of a slice over the word position in the answer. The change of the logit
relative to the beginning of the answer is shown.}\label{fig:h-position}
\end{figure}

\begin{figure}[htbp]
\centering
\includegraphics[width=\linewidth]{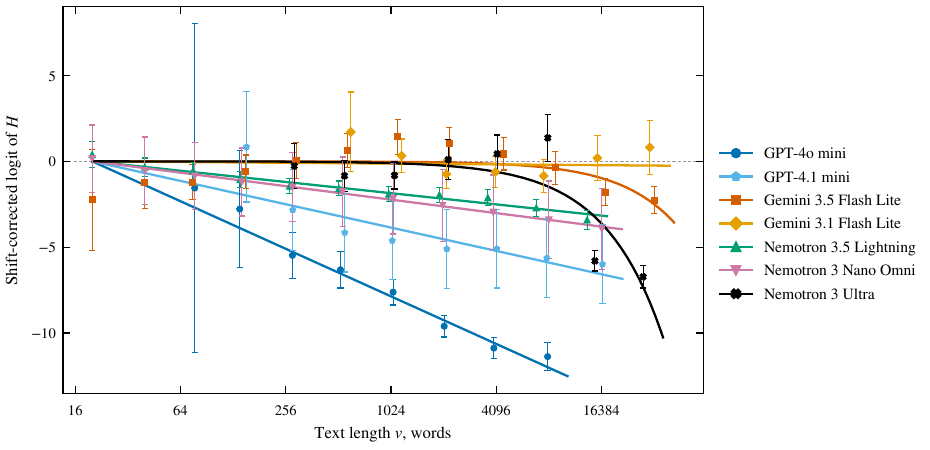}
\caption{Rule execution $H$ of seven models. Slice levels over the text length and their laws. Three undetermined levels are omitted: the
first one of Gemini 3.1 Flash Lite and the first two of Nemotron 3 Ultra.}\label{fig:h-length}
\end{figure}

\begin{figure}[htbp]
\centering
\includegraphics[width=\linewidth]{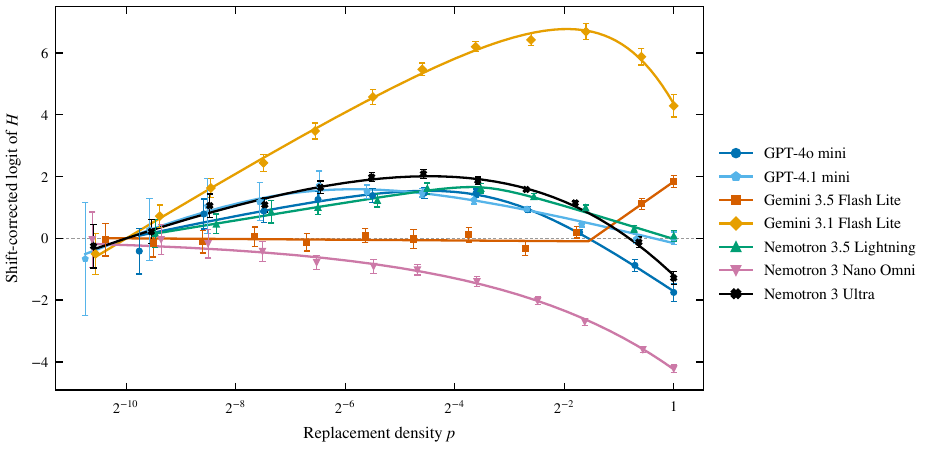}
\caption{Rule execution $H$ of seven models. Laws of the replacement density.}\label{fig:h-density}
\end{figure}

\begin{figure}[htbp]
\centering
\includegraphics[width=\linewidth]{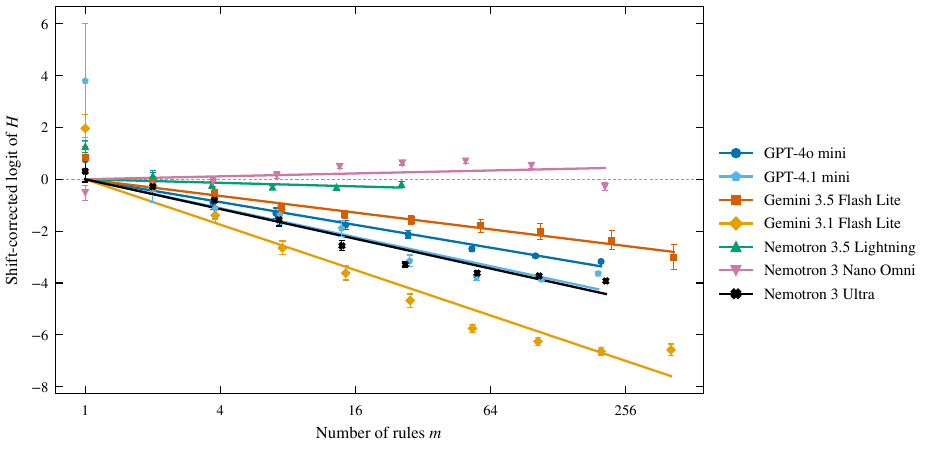}
\caption{Rule execution $H$ of seven models. Laws of the number of rules.}\label{fig:h-rules}
\end{figure}

\begin{figure}[htbp]
\centering
\includegraphics[width=\linewidth]{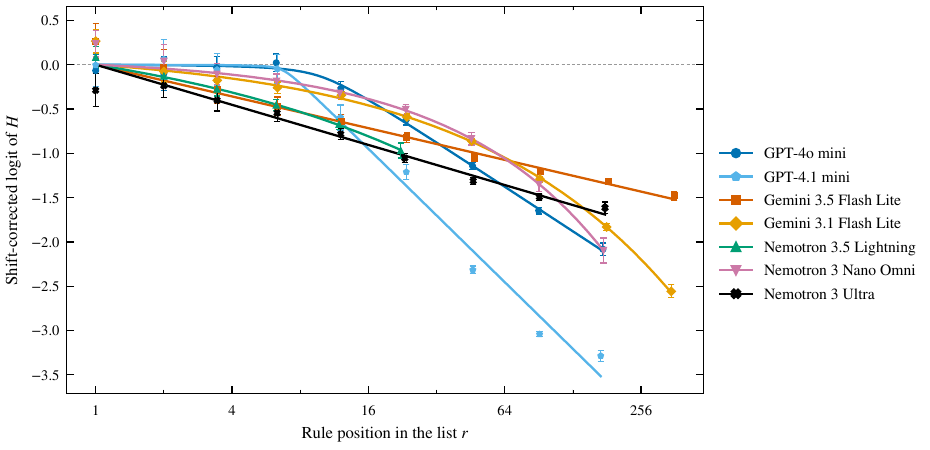}
\caption{Rule execution $H$ of seven models. Laws of the rule position in the list.}\label{fig:h-rule-position}
\end{figure}

\paragraph{Rule execution.}
The laws of rule execution are shown in Figures~\ref{fig:h-position}--\ref{fig:h-rule-position}. In all models the probability of a
correct replacement decreases along the answer and with the rule position in the list. As the text length grows, this probability
decreases in six models and hardly changes in Gemini 3.1 Flash Lite. As the number of rules grows, it decreases in five models and hardly
changes within the measured region in Nemotron 3.5 Lightning and Nemotron 3 Nano Omni.

For the replacement density the two-slope hyperbola is chosen in six models, and the binary exponential in Nemotron 3 Nano Omni. Along this
axis the differences between the models are the most pronounced.

\begin{figure}[htbp]
\centering
\includegraphics[width=\linewidth]{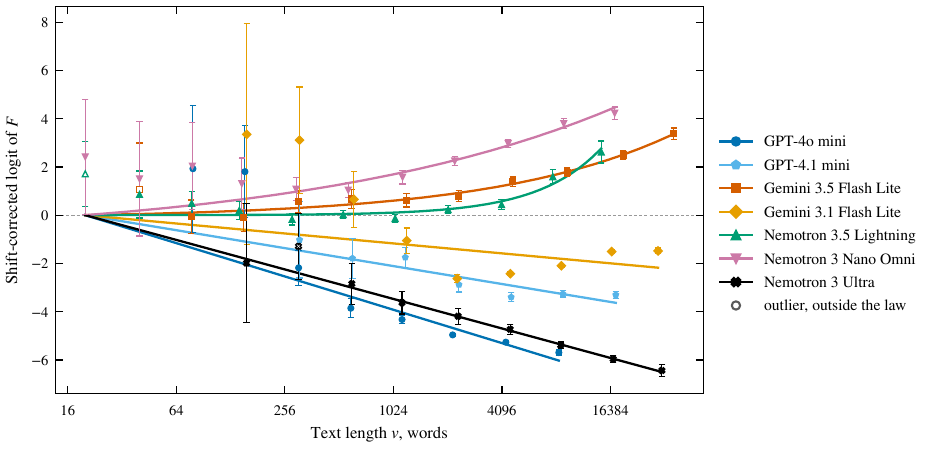}
\caption{Misfires $F$ of seven models. Laws of the text length.}\label{fig:f-length}
\end{figure}

\begin{figure}[htbp]
\centering
\includegraphics[width=\linewidth]{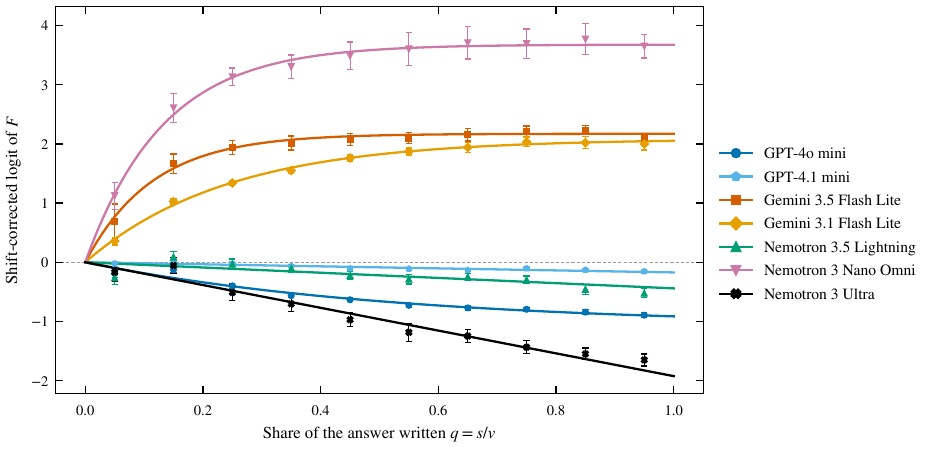}
\caption{Misfires $F$ of seven models. Laws of the share of the answer written.}\label{fig:f-answer-share}
\end{figure}

\begin{figure}[htbp]
\centering
\includegraphics[width=\linewidth]{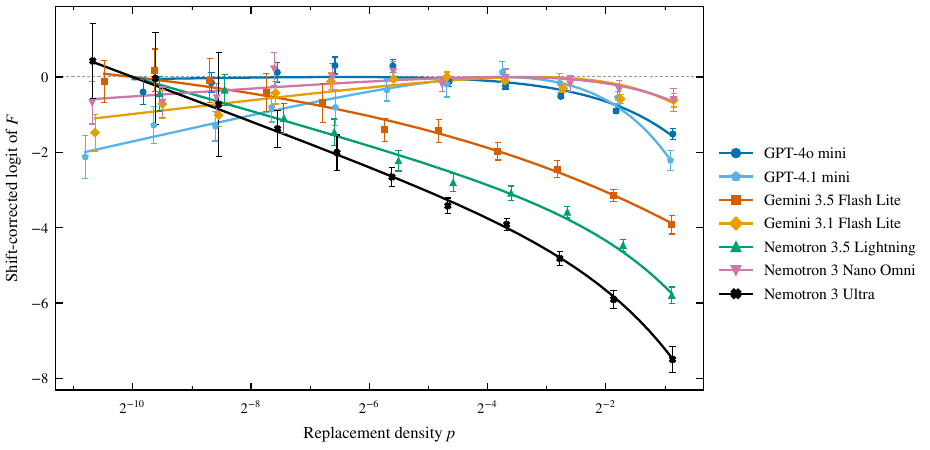}
\caption{Misfires $F$ of seven models. Laws of the replacement density.}\label{fig:f-density}
\end{figure}

\begin{figure}[htbp]
\centering
\includegraphics[width=\linewidth]{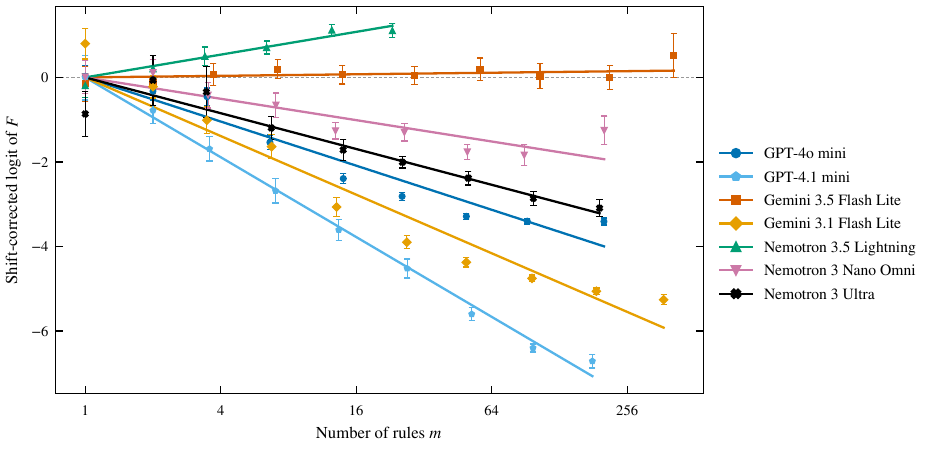}
\caption{Misfires $F$ of seven models. Laws of the number of rules.}\label{fig:f-rules}
\end{figure}

\paragraph{Misfires.}
The laws of misfires are shown in Figures~\ref{fig:f-length}--\ref{fig:f-rules}. The logit corresponds to the probability that no misfire
occurs, so a decreasing law means that misfires become more frequent.

As the replacement density grows, misfires become more frequent in all models. As the text length increases, they become more frequent in
four models and rarer in Gemini 3.5 Flash Lite, Nemotron 3.5 Lightning and Nemotron 3 Nano Omni. Towards the end of the answer misfires
become rarer in Gemini 3.5 Flash Lite, Gemini 3.1 Flash Lite and Nemotron 3 Nano Omni.

\subsection{Capacities and the measure of the density}\label{sec:passport}\label{sec:passports-density}
The measure of the density equals $e^{-\hat P(1)}$, where $\hat P(1)$ is the difference between the value of the law of the density at
$p = 1$ and its value at the zero point of the axis. The measure shows how many times the odds of an error change in the transition from
the rarest replacements to the full density. A value below one corresponds to a decrease of the odds of an error. The capacities of
misfires lie far beyond the measured region in all models or are undefined, so for $F$ only the measure of the density is given in
Table~\ref{tab:capacities}.

\begin{table}[htbp]
\centering
\scriptsize
\caption{Capacities at the level 0.5 and measures of the density of seven models. The instruction capacity is the number of rules at which
the probability of correct execution drops to 0.5. The fatigue capacity is the length of the answer in words at which synchronization or
rule execution drops to 0.5. The measure of the density is the ratio of the odds of an error at full density to the odds at the rarest
replacements. The 95\% confidence interval is given in parentheses. An asterisk marks estimates obtained by continuing a law beyond the
measured region. A dash means that the capacity is undefined, since the law does not drop to the level 0.5.}\label{tab:capacities}
\resizebox{\linewidth}{!}{%
\begin{tabular}{@{}lcccccc@{}}
\toprule
 & Instruction & \multicolumn{2}{c}{Fatigue capacity, words} & \multicolumn{3}{c}{Measure of the density} \\
\cmidrule(lr){3-4}\cmidrule(l){5-7}
Model & capacity, rules & $\Psync$ & $H$ & $\Psync$ & $H$ & $F$ \\
\midrule
GPT-4o mini & \shortstack{36.4\\(32.8--40.4)} & \shortstack{13\,515*\\(10\,993--16\,617)} & \shortstack{18\,382*\\(13\,967--24\,193)}
 & \shortstack{$1.6\cdot 10^{8}$\\($6.1\cdot 10^{7}$--$4.4\cdot 10^{8}$)} & \shortstack{5.6\\(3.0--10.3)} & \shortstack{19\\(13--27)} \\[4pt]
GPT-4.1 mini & \shortstack{23.5\\(20.9--26.4)} & \shortstack{53\,647*\\(42\,518--67\,689)} & \shortstack{169\,628*\\(93\,307--308\,378)}
 & \shortstack{$5.7\cdot 10^{12}$\\($1.6\cdot 10^{12}$--$2.0\cdot 10^{13}$)} & \shortstack{1.2\\(0.38--3.6)} & \shortstack{31\\(16--60)} \\[4pt]
Gemini 3.5 Flash Lite & \shortstack{14.6\\(13.4--15.9)} & --- & \shortstack{932\\(777--1119)}
 & \shortstack{100\\(62--163)} & \shortstack{0.16\\(0.105--0.24)} & \shortstack{325\\(82--1283)} \\[4pt]
Gemini 3.1 Flash Lite & \shortstack{18.8\\(15.3--23.2)} & \shortstack{61\,997*\\(33\,785--113\,771)} & \shortstack{4524\\(4150--4931)}
 & \shortstack{$1.5\cdot 10^{11}$\\($2.1\cdot 10^{7}$--$1.0\cdot 10^{15}$)} & \shortstack{0.013\\(0.008--0.022)} & \shortstack{2.0\\(1.3--3.0)} \\[4pt]
Nemotron 3.5 Lightning & \shortstack{10.0\\(8.0--12.6)} & \shortstack{8758\\(6532--11\,743)} & \shortstack{197\\(145--266)}
 & \shortstack{16\,861\\(4181--67\,997)} & \shortstack{1.0\\(0.66--1.6)} & \shortstack{1933\\(1022--3657)} \\[4pt]
Nemotron 3 Nano Omni & \shortstack{34.3\\(29.8--39.4)} & \shortstack{29\,156*\\(12\,520--67\,900)} & \shortstack{842\\(716--990)}
 & \shortstack{$2.8\cdot 10^{7}$\\($4.3\cdot 10^{5}$--$1.8\cdot 10^{9}$)} & \shortstack{69\\(45--106)} & \shortstack{2.6\\(1.3--5.3)} \\[4pt]
Nemotron 3 Ultra & \shortstack{17.0\\(15.8--18.1)} & \shortstack{84\,432*\\(58\,255--122\,372)} & \shortstack{39\,459\\(35\,258--44\,161)}
 & \shortstack{$1.3\cdot 10^{10}$\\($2.5\cdot 10^{9}$--$6.2\cdot 10^{10}$)} & \shortstack{3.3\\(2.1--5.2)} & \shortstack{17\,479\\(6907--44\,234)} \\
\bottomrule
\end{tabular}%
}
\end{table}

\subsection{Features of the behavior of the models}\label{sec:passports-features}\label{sec:passports-rises}
In all seven models synchronization worsens along the answer and as the replacement density grows, and rule execution worsens along the
answer and with the rule position in the list. Misfires become more frequent as the density grows. The most pronounced differences between
the models are related to the replacement density. At full density the odds of losing synchronization grow by a factor from $100$ to
$5.7\cdot 10^{12}$ depending on the model. The odds of a wrong replacement grow in some models and decrease in others. On some axes the
quality improves as the load grows. These segments are shown in the passports, and the explanation of their origin is beyond the scope of
this paper.

GPT-4o mini has the largest estimate of the instruction capacity, 36.4 rules. As the text length grows, rule execution in this model
worsens the most. Misfires become more frequent both as the text length grows and as the number of rules increases.

GPT-4.1 mini keeps rule execution along the answer longer than the other models. Its fatigue capacity of $H$ is 169\,628* words. The
replacement density has little effect on rule execution. At the same time, synchronization in this model depends on the replacement
density the most, with a measure of the density of $5.7\cdot 10^{12}$, and worsens noticeably as the number of rules grows.

In Gemini 3.5 Flash Lite synchronization improves as the text length increases, so the fatigue capacity of synchronization of this model
is undefined. The dependence of synchronization on the replacement density is the weakest in this model. At full density the odds of a
wrong replacement decrease sixfold. Misfires become rarer as the text length increases and towards the end of the answer.

In Gemini 3.1 Flash Lite the odds of a wrong replacement at full density decrease 78 times. Rule execution in this model worsens the most
along the answer and as the number of rules grows.

Nemotron 3.5 Lightning has the smallest instruction capacity and the smallest fatigue capacity of rule execution, 10 rules and 197 words
respectively. Within the measured region the number of rules hardly affects their execution, and synchronization improves as the number
of rules grows. Misfires do not become more frequent as the text length or the number of rules increases.

Nemotron 3 Nano Omni, with an estimated instruction capacity of 34.3 rules, is close to GPT-4o mini. The full density hinders rule
execution in this model the most. The odds of a wrong replacement grow 69 times, and the law of the density has the form of a binary
exponential. Misfires become rarer as the text length increases and towards the end of the answer.

Nemotron 3 Ultra keeps synchronization along the answer longer than the other models. Its fatigue capacity of $\Psync$ is 84\,432*
words. By the fatigue capacity of rule execution, 39\,459 words, it ranks second. The dependence of misfires on the replacement density is
the most pronounced in this model, with a measure of the density of 17\,479. Misfires also become more frequent as the text length
increases.

\section{Outlook}\label{sec:outlook}

The laws of the passport describe the coarse structure of the behavior of a model---the dependence of the probability of an outcome,
averaged over requests and positions, on each kind of load. These laws make it possible to estimate the size and the rate of the decline
in quality. At the level of an individual word the subject of study becomes the fine structure of behavior---the dependence of the outcome
on the immediate surroundings of the word in the source text and in the answer. Its study is the next stage of the research.

The first direction is the influence of the proximity of words on the outcome of replacing or keeping a word. It is taken into account
whether the neighboring words are subject to replacement and whether they are similar to the word under consideration. The second
direction is the influence of the position of a word within a line, taking into account its position in the answer as a whole. The third
is the mechanism of copying the source text and the conditions under which copying breaks down. The fourth is the mechanism of
substitution. It remains to find out which word the model puts in place of the correct one when it errs and how an erroneous replacement
of a background word differs from an error in a rule word.

The alignment of the answer with the reference makes it possible to determine the outcome of every position and to relate it to the
surroundings of the word in the source text and in the answer. The analysis of the fine structure will reveal features of the behavior of
a neural network that are smoothed out by averaging, including the segments where quality grows with the load. The instrument measures
the form of a dependence. The mechanisms proposed to explain it require verification in separate experiments.

\section{Code and Data}\label{sec:availability}

{\sloppy
The public part of the instrument is published in the open repository \url{https://github.com/kutomanovhennadii/instruction-capacity},
and this work describes its version 0.2.3. The repository contains the code that executes one point, the code that aligns the answer with
the reference, the code that builds the passport of the three quantities and checks it, the vocabularies together with the code that
makes them, the documents of the method and the passports of all seven models compared in Section~\ref{sec:passports}. The passports of
the seven models are also available in a collection of datasets on Hugging Face
\url{https://huggingface.co/collections/kutomanovhennadii/iec-passports-6ae3ad9fe2c798814eb8a3dc}.\par}

The measured points are published only for the models whose terms of access allow it. The answers of GPT-4o mini and GPT-4.1 mini were
obtained through the OpenAI API, and under OpenAI's terms for API customers the output belongs to the customer (OpenAI Services
Agreement, Section~4.1). The answers of Gemini 3.5 Flash Lite and Gemini 3.1 Flash Lite were obtained through the Gemini API, and under the
Gemini API Additional Terms of Service Google does not claim ownership of the generated content. The measurements of these four models are
published in full: every measurement contains the records of all points with the prompts, answers and token counts, a description of the
measurement with checksums of the files, and the passport. The answers of Nemotron 3.5 Lightning, Nemotron 3 Nano Omni and Nemotron 3
Ultra were obtained through the free endpoints that NVIDIA provides on OpenRouter, and the NVIDIA API Trial Terms of Service
(Section~4.2) do not allow the generated content to be distributed. Therefore the measured points of these three models are not
published: only their passports are published, which contain statistics of the answers but not the answers themselves, and they cannot
be recomputed from points.

Any number of the passports of the four fully published models given in the paper is recomputed from the published points. For GPT-4o
mini this is done by the command
\begin{verbatim}
python scripts/build_passport.py artefacts/pilot_gpt4o_mini_v55 \
    --out-dir my_passport
\end{verbatim}
and for the other models the command takes the directories
\begin{verbatim}
artefacts/pilot_gpt4_1_mini_v56
artefacts/pilot_gemini3_5_flash_lite_v58
artefacts/pilot_gemini3_1_flash_lite_v57
\end{verbatim}
The integrity and reproducibility of a published measurement are checked by the command
\begin{verbatim}
python scripts/verify.py artefacts/pilot_gpt4o_mini_v55
\end{verbatim}
The check verifies the checksums of the files, compares the vocabulary with its reference version, regenerates the list of rules and the
text of every point from its seeds and recomputes the passport, comparing it with the published one within the computational
precision; it does not query the language model. The same code can be used to measure points on another model and to build its
passport.

The numbers that the paper gives for the measurement of the previous work \citep{kutomanov2026capacity} refer to its 2958 requests to
GPT-4o mini. This measurement is stored in the same repository under the tag \texttt{v0.1.0}, and its points are part of the published
GPT-4o mini measurement.

The points of the published measurements were chosen by an adaptive planning technology developed by the author: it decides which points
to measure from the results of those already measured. This technology forms the closed part of the instrument, and neither its design
nor the rules for choosing points are published. The passports do not depend on it: they are computed only from the published points,
regardless of how these were chosen, so every number of a passport can be checked without the closed part.

The code is distributed under the MIT license, the data and documents under the Creative Commons Attribution 4.0 International license.

\section*{Acknowledgments and Use of AI}

The author thanks Dipankar Sarkar for an independent recomputation of the published points of the previous work
\citep{sarkar2026comment}: the recomputation revealed replacements of background words that the former outcome did not account for, and
the change of the method described in Section~\ref{sec:three-quantities} began with this question.

In developing the code of the instrument and the statistical models, in processing the data and in preparing the text of the paper, the
author used Claude models by Anthropic. The formulation of the problem, the design of the instrument and the method, and the decisions on
the content and form of the results belong to the author; the author has checked and is responsible for the entire content of the paper,
including the code and the numbers.

The measurements were carried out on free access provided by the developers of the models and by platforms. The author thanks OpenAI for
the program of complimentary daily tokens for traffic shared with OpenAI\footnote{\url{https://help.openai.com/en/articles/10306912}.}:
organizations that allow their API traffic to be shared with OpenAI receive up to 10~million free tokens per day for small models,
including gpt-4o-mini and gpt-4.1-mini, and the measurements of GPT-4o mini and GPT-4.1 mini were carried out on these tokens. The author
thanks Google for the free tier of the Gemini API, on which the measurements of Gemini 3.5 Flash Lite and Gemini 3.1 Flash Lite were
carried out, and NVIDIA and OpenRouter for the free endpoints of the Nemotron models, on which the measurements of the three Nemotron models
were carried out.

\bibliographystyle{plainnat}
\bibliography{references}
\end{document}